\documentclass[lettersize,journal]{IEEEtran}

\usepackage{acro}
\usepackage{multirow}
\usepackage{xcolor}

\usepackage{enumitem}
\usepackage{authblk}

\usepackage{graphics}
\usepackage{graphicx} 
\usepackage{multirow}
\usepackage{longtable}
\usepackage{epsfig}
\usepackage{epstopdf}

\usepackage{amsmath, mathrsfs, dsfont}
\usepackage{amssymb}
\usepackage{amsfonts}
\usepackage{amsbsy}
\usepackage{bm}

\usepackage{cite}
\usepackage{verbatim}

\usepackage{amscd}
\usepackage{color}
\definecolor{lgray}{gray}{0.6}
\usepackage{rotating}

\usepackage[font=footnotesize]{subcaption}

\usepackage{mathptmx}
\usepackage[11pt]{moresize}
\usepackage{flushend}
\usepackage[stable]{footmisc}

\usepackage{algorithmicx}
\usepackage{algorithm}
\usepackage{algpascal}
\usepackage{float}
\usepackage{algc}
\usepackage{algcompatible}
\usepackage{algpseudocode}

\usepackage[vcentermath]{youngtab} 
\usepackage{url}

\newcommand{\BoldA}				{ \mathbf{A} }

\newcommand{\BoldB}				{ \mathbf{B} }

\newcommand{\BoldC}				{ \mathbf{C} }

\newcommand{\Bolde}				{ \mathbf{e} }

\newcommand{\Boldf}				{ \mathbf{f} }

\newcommand{\BoldH}				{ \mathbf{H} }

\newcommand{\BoldI}				{ \mathbf{I} }
\newcommand{\I}					{ \mathbf{I} }

\newcommand{\BoldK}				{ \mathbf{K} }

\newcommand{\Boldp}				{ \mathbf{p} }
\newcommand{\BoldQ}				{ \mathbf{Q} }

\newcommand{\BoldR}				{ \mathbf{R} }

\newcommand{\Bolds}				{ \mathbf{s} }

\newcommand{\BoldT}				{ \mathbf{T} }

\newcommand{\Boldv}				{ \mathbf{v} }

\newcommand{\BoldX}				{ \mathbf{X} }

\newcommand{\Boldy}				{ \mathbf{y} }
\newcommand{\BoldZ}				{ \mathbf{Z} }
\newcommand{\Boldz}				{ \mathbf{z} }

\newcommand{\0}					{ \boldsymbol{0} }

\newcommand{\Boldomega}			{ \boldsymbol{\omega} }
\newcommand{\Boldnu}			{ \boldsymbol{\nu} }

\newcommand{\BoldPhi}			{ \boldsymbol{\Phi} }
\newcommand{\Boldphi}			{ \boldsymbol{\phi} }
\newcommand{\Boldpi}			{ \boldsymbol{\pi} }

\newcommand{\E}					{\mathrm{E}}

\newcommand\T{\rule{0pt}{2.6ex}}       
\newcommand\B{\rule[-1.2ex]{0pt}{0pt}} 

\newcounter{inlineenum}
\renewcommand{\theinlineenum}{\alph{inlineenum}}

\DeclareMathAlphabet{\mathcal}{OMS}{cmsy}{m}{n}

	\newcommand{\brmrk}[1]{\begin{remark} \label{#1} }
	\newcommand{\ermrk}{ \hfill $\bigtriangleup$    \end{remark} \vspace{1mm} }

\newtheorem{exercise}{Exercise}[section]
\newcommand{\boex}[1]{\begin{example} \label{#1} --- \rm}
	\newcommand{\eoex}{ \hfill $\bigtriangleup$    \end{example} \vspace{1mm} }
\newtheorem{example}{Example}[section]
\newcommand{\bohw}[1]{\begin{exercise} \label{#1} -- \rm}
	\newcommand{\eohw}{ \hfill    \end{exercise} \vspace{1mm} }
\newtheorem{assumption}{Assumption}[section]
	\newcommand{\boass}[1]{\begin{assumption} \label{#1} -- \rm}
	\newcommand{\eoass}{ \hfill    \end{assumption} \vspace{1mm} }

\newcommand{\todo}[1]{\footnote{\color{green}TO DO: {#1}}}

\definecolor{brinkpink}{rgb}{1.00, 0.33, 0.64}



\DeclareAcronym{APC}{
	short=APC,
	long=Antenna Phase Centre,
}

\DeclareAcronym{COM}{
short=COM,
long=Center of Mass,
}

\DeclareAcronym{CV}{
short=CV,
long=Connected Vehicle,
}

\DeclareAcronym{CAV}{
	short=CAV,
	long=Connected and Automated Vehicle
}

\DeclareAcronym{DCB}{
short=DCB,
long=Differential Code Bias,
}

\DeclareAcronym{DSRC}{
	short=DSRC,
	long=Dedicated Short-Range Communications,
}

\DeclareAcronym{CDMA}{
	short=CDMA,
	long=Code Dvision Multiple Access,
}

\DeclareAcronym{CDF}{
	short=CDF,
	long=Cumulative Distribution Function,
}

\DeclareAcronym{CE-CERT}{
	short=CE-CERT,
	long=College of Engineering Center for Environmental Research and Technology,
}

\DeclareAcronym{FDMA}{
	short=FDMA,
	long=Frequency Dvision Multiple Access,
}
\DeclareAcronym{DGPS}{
	short=DGPS,
	long=Differential Global Positioning System,
}
\DeclareAcronym{DGNSS}{
short=DGNSS,
long=Differential GNSS,
}
\DeclareAcronym{ECEF}{
short=ECEF,
long=Earth-Centered Earth-Fixed,
}
\DeclareAcronym{GPS}{
	short=GPS,
	long=Global Positioning System,
}
\DeclareAcronym{GNSS}{
short=GNSS,
long=Global Navigation Satellite System,
}
\DeclareAcronym{GPSSPS}{
short=GPS SPS,
long=GPS standard positioning service,
}

\DeclareAcronym{HD}
{short = HD, long = Horizontal Distance}

\DeclareAcronym{HiD}{
	short=Hi-Def,
	long=High-definition,
}

\DeclareAcronym{IMU}{
short=IMU,
long=Inertial Measurement Unit,
}

\DeclareAcronym{INS}{
	short=INS,
	long=Inertial Navigation System,
}

\DeclareAcronym{I2V}{
short=I2V,
long=Infrastructure-to-Vehicle,
}

\DeclareAcronym{V2X}{
short=V2X,
long=Vehicle-to-Everything
}

\DeclareAcronym{LLMM}
{short = LLMM,
	 long = Lane-level Map-matching}

\DeclareAcronym{LD}
{short = LD,
	 long = Lane Determination}

\DeclareAcronym{RLMM}
{short = RLMM,
	 long = Road-level Map-matching}

\DeclareAcronym{MSE}
{short = MSE,
	 long = Mean Square Error}

\DeclareAcronym{PDF}{
	short=PDF,
	long=Probability Distribution Function,
}

\DeclareAcronym{OS}{
	short=OS,
	long=Open Service,
}

\DeclareAcronym{OSB}{
short=OSB,
long=Observable-specific Code Biases,
}
\DeclareAcronym{OSR}{
short=OSR,
long=Observation Space Representation,
}
\DeclareAcronym{PPP}{
	short=PPP,
	long=Precise Point Positioning,
} 
\DeclareAcronym{PPP-AR}{
short=PPP-AR,
long=Precise Point Positioning Ambiguity Resolution,
} 
\DeclareAcronym{RTCM}{
short=RTCM,
long=Radio Technical Commission for Maritime Services,
}
\DeclareAcronym{RTK}{
short=RTK,
long=Real-time Kinematic Positioning,
}
\DeclareAcronym{SBAS}{
short=SBAS,
long=Satellite Based Augmentation Systems,
}

\DeclareAcronym{SNR}{
short=SNR,
long=Signal-to-Noise Ratio,
}
\DeclareAcronym{SSR}{
short=SSR,
long=State Space Representation,
}
\DeclareAcronym{SPS}{
	short=SPS,
	long=Standard Positioning Service,
}

\DeclareAcronym{SAE}{
	short=SAE,
	long=Society of Automotive Engineers,
}
\DeclareAcronym{STEC}{
	short=STEC,
	long=Slant Total Electron Content,
}
\DeclareAcronym{VTEC}{
	short=VTEC,
	long=Vertical Total Electron Content,
}

\DeclareAcronym{SH}{
	short=SH,
	long=Spherical Harmonic,
}
\DeclareAcronym{TGD}{
short=TGD,
long=Timing Goup Delay,
}
\DeclareAcronym{ZTD}{
short=ZTD,
long=Zenith Troposphere Delay,
}
\DeclareAcronym{TEC}{
short=TEC,
long=Total Electron Content,
}
\DeclareAcronym{IPP}{
	short=IPP,
	long=Ionosphere Pierce Point,
}
\DeclareAcronym{NOAA}{
	short=NOAA,
	long=National Oceanic and Atmospheric Administration,
}
\DeclareAcronym{UCR}{
	short=UCR,
	long=University of California-Riverside,
}
\DeclareAcronym{USTEC}{
short=US-TEC,
long=US Total Electron Content,
}
\DeclareAcronym{VNDGNSS}{
	short=VN-DGNSS,
	long=Virtual Network DGNSS,
}
\DeclareAcronym{IOD}{
	short=IOD,
	long=Issue Of Data,
}
\DeclareAcronym{CAS}{
	short=CAS,
	long=Chinese Academy of Sciences,
}
\DeclareAcronym{CNES}{
	short=CNES,
	long=Centre national d'études spatiales,
}
\DeclareAcronym{RMS}{
	short=RMS,
	long=Root Mean Square,
}
\DeclareAcronym{SF}{
	short=SF,
	long=Single Frequency,
}
\DeclareAcronym{STD}{
	short=STD,
	long=Standard Deviation,
}
\DeclareAcronym{DF}{
	short=DF,
	long=Dual Frequency,
}
\DeclareAcronym{ICD}{
	short=ICD,
	long=Interface Control Document,
}
\DeclareAcronym{NED}{
	short=NED,
	long={North, East, and Down},
}
\DeclareAcronym{WAAS}{
	short=WAAS,
	long=Wide Area Augmentation System,
}
\DeclareAcronym{OPUS}{
	short=OPUS,
	long=Online Positioning User Service,
}

\DeclareAcronym{VRS}{
	short=VRS,
	long=Virtual Reference Station,
}

\DeclareAcronym{SPP}{
	short=SPP,
	long=Single-frequency Point Positioning,
}

\DeclareAcronym{SPaT}{
	short=SPaT,
	long=Signal Phase and Timing,
}

\DeclareAcronym{BNC}{
	short=BNC,
	long=BKG NTRIP Client,
}

\DeclareAcronym{ITS}{
	short=ITS,
	long=Intelligent Transportation Systems
}

\DeclareAcronym{HMM}{
	short = HMM, 
	long = Hidden Markov Model
}

\DeclareAcronym{USDOT}{
	short=USDOT,
	long=U.S. Department of Transportation
}

\begin{document}
	\title{Data-Dependent Hidden Markov Model with Off-Road State Determination and Real-Time Viterbi Algorithm for Lane Determination in Autonomous Vehicles}
\author{Mike~Stas,
			Wang~Hu,
		and~Jay~A.~Farrell
	\thanks{This work is intended for submission to the IEEE for possible publication. Copyright may be transferred without notice, after which this version may no longer be accessible.}
}
\affil{Department of Electrical and Computer Engineering, University of California, Riverside, CA, USA}
\affil[]{Email: \texttt{mstas001@ucr.edu}, \texttt{whu027@ucr.edu}, \texttt{farrell@ece.ucr.edu}}

	\maketitle
	
	\begin{abstract}
	 	Lane determination and lane sequence determination are important components for many \ac{CAV} applications.
	 	Lane determination has been solved using \ac{HMM} among other methods.
	 	The existing \ac{HMM} literature for lane sequence determination
	    uses empirical definitions with user-modified parameters to calculate \ac{HMM} probabilities.
	 	The probability definitions in the literature can cause breaks in the \ac{HMM}  due to the inability to directly calculate probabilities of off-road positions, requiring post-processing of data. 
	 	This paper develops a time-varying \ac{HMM} using the physical properties  of  the roadway and vehicle, and the stochastic properties of the sensors.
	 	This approach yields emission and transition probability models conditioned on the sensor data without parameter tuning.
	 	It also accounts for the probability that the vehicle is not in any roadway lane (e.g., on the shoulder or making a U-turn), which eliminates the need for post-processing to deal with breaks in the \ac{HMM} processing.

	 	This approach requires adapting the Viterbi algorithm and the \ac{HMM} to be conditioned on the sensor data, which are then used to generate the most-likely sequence of lanes the vehicle has traveled.
	 	The proposed approach achieves an average accuracy of 
	 	$95.9\%$.  
	 	Compared to the existing literature, this provides an average increase of 
	 	$2.25\%$ by implementing the proposed  transition probability and an average increase of 
	 	$5.1\%$ by implementing both the proposed transition and emission probabilities.
	\end{abstract}
\begin{IEEEkeywords}
	 Lane-level map matching, Map matching, Hidden Markov Models (HMM), Viterbi algorithm, Real-time Viterbi algorithm, State estimation 
\end{IEEEkeywords}

\section{Introduction} \label{sec:Intro}

\IEEEPARstart{A}{ccurate} vehicle lane determination is essential to \ac{ITS} applications  that require  knowledge of the vehicle's current lane for interactions between a vehicle and its neighbors or the roadway infrastructure, for example,  \ac{V2X} \ac{CAV} systems,  analysis of lane-level vehicle trajectories, traffic patterns, platoon trajectory planning, and intersection management \cite{lanelocal2023,altan2017glidepath}.
With the enhanced accuracy of real-time vehicle navigation systems and high-definition maps \cite{hu2022assessment}, lane determination is becoming feasible via lane-level map matching  \cite{du2004lane,du2008next,rabe2016lane,li2017lane}.
Lane-level map matching methods differ in how they approach matching estimated vehicle positions to maps \cite{du2004lane,du2008next,rabe2016lane,li2017lane,Goh2012,Hu2019,Atia2017,Fu2019,Hannson2021}.

Map matching methods can be split into three generic approaches.
First, point-to-point map matching where a vehicle position is being matched to a node on the map.
Second, point-to-curve map matching where a vehicle position is being matched to a lane line (e.g. center line) on the map.
Finally, curve-to-curve map matching where a vehicle trajectory is being matched to a lane line (e.g. center line) on the map. 
The methods described in \cite{du2004lane,du2008next} use a curve-to-curve map matching approach to match vehicle positions to a lane map.
The method described in \cite{rabe2016lane} solves the lane determination problem through least squares optimization, while the method described in \cite{li2017lane} uses a particle filter.
Articles \cite{Goh2012,Hu2019,Atia2017,Fu2019,Hannson2021,Newson} utilize \ac{HMM} to solve map matching problems.

Before navigation systems were accurate enough to enable lane-determination, the
\ac{HMM} was studied for road-level map matching in \cite{Goh2012,Hu2019,Newson}.
This work was extended by  \cite{Atia2017} and \cite{Fu2019} to demonstrate lane-level map matching.
Their \ac{HMM} used the vehicle position, velocity, and other measurements to generate emission and transition probabilities for road-level map matching.
Once the road was determined, the position error relative to each lane center-line was used for lane-determination.

More recently, \cite{Hannson2021} used a collection of sensor  data  such as GPS observations, speed,  heading,  turn signal, and marker type variables to calculate factors that were multiplied, then normalized to compute an emission probability model.
To calculate the transition probability, a maximum depth parameter alongside an added tuning parameter was used, then normalized, to generate the required lane transition probability matrix.


While the existing literature uses the \ac{HMM} and the Viterbi algorithm to provide working lane-determination methods, there are two main literature gaps that need to be addressed:
\begin{itemize}
	\item There exists no method of calculating the probability of the vehicle being off-road which causes \ac{HMM} breaks when the vehicle is not in any lane, for example in an intersection.
	The literature deals with such breaks through post processing by removing off-road positions from the dataset.
	Even though post-processing can deal with \ac{HMM} breaks, the break might be too long where \ac{HMM} re-initialization is required, thus losing previous vital observation information.
	Moreover there is no real-time \ac{HMM} approach that can incorporate off-road positions into the model without re-initializing the \ac{HMM} at every break.
	\item \ac{HMM} transition probability definitions do not condition on observations, instead rely on ad-hoc definitions.
	While such definitions work, they often depend on fixed probabilities or parameter tuning to improve performance on tested datasets which limits their generalizability as the \ac{HMM} probabilities will need to be tuned in post processing.
	This reliance in tuning, which is often overlooked or not thoroughly explained, makes integration  into existing \ac{CAV} systems more challenging.
\end{itemize}

The paper aims at addressing the gaps by introducing a data-dependent \ac{HMM} model and a real-time Viterbi algorithm.
The contributions of this paper are:
\begin{itemize}
	\item Presenting an \ac{HMM} model that  directly accounts for the probability of the vehicle being off-road, thus eliminating the possibility of any \ac{HMM} breaks.
	\item Developing time-varying and data-dependent emission and transition probability models built on state-space and stochastic models of the vehicle and its navigation systems which are interpreted within the context of lane geometry described by a high-definition roadway map.
	\item Presenting batch and real-time Viterbi lane determination algorithms that incorporate the proposed data-dependent \ac{HMM}.
	\item Proposing a real-time Viterbi lane determination algorithm that has a more informed initialization step by propagating the initial state probability from previous time windows.
	\item Demonstrating improved performance relative to the existing methods in the literature without any probability parameter tuning or post processing of data to remove off-road observations.
\end{itemize}

In \ac{HMM}, the actual lane the vehicle is in at each time is a hidden state that the model tries to infer.
Given a sequence of vehicle state vector estimates and its error covariance matrix from the navigation system, the vehicle has some probability of being in each lane at each instant.
 Then,  the sequence of lane transitions can be stochastically modeled and estimated based on the mean and covariance of the vehicle position and velocity, along with the high-definition map, to determine the probability of lane transitions at each time. 
The \ac{HMM} has emerged as a powerful tool utilizing the  emission  and state transition models to determine the most likely sequence of lanes and lane transitions (i.e., the hidden states). 

The paper is organized as follows:
 Section \ref{sec:Intro} Introduces lane determination, reviews existing related works, and presents literature gaps to be addressed. 
Section \ref{sec:HmmComp} states the problem, and reviews and extends \ac{HMM} definitions and properties  to create the data-dependent \ac{HMM} while adapting it to the lane determination problem. 
Section \ref{sec:HMM} presents the Viterbi algorithm to determine the most likely sequence of lanes, in the context of lane determination where the HMM is time-varying and data-dependent.
It also  proposes a real-time approach to solving the Viterbi algorithm. 
Section \ref{sec:LaneFrame} presents the lane model used throughout the paper.
Section \ref{sec:EmiP} presents the proposed data-dependent emission probability model.
Section \ref{sec:TranP} presents the proposed data-dependent transition probability model.
Section \ref{sec:NP_Lanes} presents the required model modifications for non-parallel lane scenarios.
Section \ref{sec:LD_model} describes the method of selecting the road segment.
Section \ref{sec:Analysis} introduces the experimental set-up and procedure, then analyzes the results and performance of different models.
Finally, Section \ref{sec:Conc} concludes the paper and presents future work.


\section{Lane Determination Hidden Markov Model }  \label{sec:HmmComp}
\noindent
This section reviews \ac{HMM}s while adapting their definition \cite{jurafsky_martin,Rabiner} to the lane determination problem  by introducing  data-dependent \ac{HMM} .\label{subsec:lambda}
The complete \ac{HMM} is denoted
\begin{align}
	\lambda = (\BoldA_k,\,\BoldB_k,\,\Boldpi^+) ,~k = 0,\ldots,K. \label{eqn:lambda}
\end{align}

Refer to Table \ref{Table:HMM} for symbols used throughout this section.

The symbols $\BoldA_k$, $\BoldB_k$, $\Boldpi^+$ and $K$ will be defined in the following subsections. 

\begin{table}[b]
	\centering
	\begin{tabular}{|p{1cm}||p{6cm}|}
		\hline
		Symbol &Definition \\
		\hline\hline
		$\lambda$ & Complete Hidden Markov Model \\
		\hline
		$L_i $ & Lane $i$ in the set of available lanes $\mathcal{L}_k$\\
		\hline
		$\pi_i^-, \pi_i^+$ & Prior and posterior initial state probability of starting in Lane $i$ \\
		\hline
		$O_k$ & Observation at time $t_k$ \\
		\hline
		$q_k $ & The hidden state corresponding to the lane in which the vehicle is located at time $t_k$ \\
		\hline
		$\BoldX_{k} $ & The vehicle state vector at time $t_k$ \\
		\hline
		$\hat\BoldX_{k}^-, \hat\BoldX_{k}^+$ & Prior and posterior vehicle state estimates at time $t_k$\\
		\hline
		$\BoldC_{\BoldX_k}^-, \BoldC_{\BoldX_k}^+$ & Prior and posterior covariance matrix at time $t_k$\\
		\hline
		$b_i(O_k) $ & Emission probability of observing $O_k$ given that the system is in lane $L_i$ \\
		\hline
		$a_{ij}^k $ & The probability of the lane state transitioning from $L_i$ at time $t_k$ to $L_j$ at time $t_{k+1}$\\
		\hline

	\end{tabular}
	\caption{Important symbol definitions used throughout Section \ref{sec:HmmComp}}
	\label{Table:HMM}
\end{table}

\subsection{Problem Statement}
Given a digital map of multi-lane road segments and a sequence of observations $O_k$ for times $t_k=k\,T$ with $k=0,\ldots,K,$  where $K$ is the total number of observations  and the sample time $T$ being positive, the challenge is to find the most likely sequence of lanes occupied by the vehicle at each time in this interval. 
Each observation $O_k$ is an estimate of the mean vehicle position  and velocity along with their error covariance matrix.
 At each time $t_k$, the road segment is defined by a set of $N$ lanes, indexed by $i$ 
\begin{align}\label{eqn:laneset_def}
	\mathcal{L}_k=\{L_i\}_{i=0}^N.
\end{align}
The lanes $L_i$ for $i\in\{1,\ldots,N\}$ are the physical lanes of the road segment. 
The lane $L_0$ accounts for the vehicle being off the road (i.e., not in a physical lane).
At each time $t_k$, the vehicle is in one of $L_i\in\mathcal{L}_k$ either in any one of the physical lanes or off the road. 
Each lane's occupancy will be more or less likely based on the measurement $O_k$ and the lane state at time step $t_{k}$.
Between times $t_{k}$ and $t_{k+1}$ some lane transitions will be more or less likely, due to the physical road layout and the state vector of the vehicle.
Combining these measurements and state transition probabilities to determine the optimal state sequence is where the HMM and Viterbi algorithm are useful.

Table \ref{Table:Index} summarizes indexing variable definitions used throughout the paper.

\begin{table}[b]
	\centering
	\begin{tabular}{|p{1cm}||p{5cm}|}
		\hline
		Symbol &Definition \\
		\hline\hline
		$k$ & Time step index\\
		\hline
		$i$ & Lane index at the current time step \\
		\hline
		$j$ & Lane index at the next time step\\
		\hline
		$l$ & Lane index at a previous time step\\
		\hline
		$m$ & Road segment index \\
		\hline
		
	\end{tabular}
	\caption{Index definitions used throughout the paper}
	\label{Table:Index}
\end{table}

\subsection{Hidden State $q_k$}
The hidden state $q_k$ is the lane in which the vehicle is located at time $t_k$ with
 $q_k = L_i$ for some $i\in\{0,\ldots,N\}$ where $q_k \in \mathcal{L}_k$.

\subsection{Vehicle State-Space Model}
The vehicle state vector is defined as\footnote{Within the navigation system, the state vector may have additional components, but only position and velocity are used herein.}
$${\BoldX}_k =\begin{bmatrix}
	{\Boldp}_k &
	{\Boldv}_k
\end{bmatrix}^\top$$ 
where $\Boldp_k$ and $\Boldv_k$ are  the 2D  position and velocity.
At any time $t_k$, the vehicle navigation system provides both the prior and posterior state estimates with their error covariance matrices.
The state estimates  and covariances are defined in Chapter 5 of \cite{maybeck1979stochastic} as
\begin{align}
	\hat{\BoldX}_{k}^+ &= \E \left\langle \BoldX_{k} | ~ y_0,...,y_k \right\rangle\nonumber\\
	\BoldC_{\BoldX_k}^+ &= \E \left \langle (\BoldX_k-\hat\BoldX_k^+)(\BoldX_k-\hat\BoldX_k^+)^\top| ~ y_0,...,y_k \right\rangle\nonumber\\
	\hat{\BoldX}_{k+1}^- &= \E \left\langle \BoldX_{k+1} | ~ y_0,...,y_k \right\rangle \nonumber\\
	\BoldC_{\BoldX_{k+1}}^- &= \E \left \langle (\BoldX_{k+1}-\hat\BoldX_{k+1}^-)(\BoldX_{k+1}-\hat\BoldX_{k+1}^-)^\top| ~ y_0,...,y_k\right\rangle \nonumber
\end{align}
 where $y_k$ represent the navigation system sensor measurements at time $t_k$. 
The vehicle navigation system could, for example, include a \ac{GNSS}  \cite{hofmann2007gnss,teunissen2017springer,hu2022using}, a \ac{GNSS} aided \ac{INS}  \cite{kreibich2021lane,rahman2022low},  or any other localization system used  to determine the vehicle state.
The methods discussed herein apply regardless of which sensors are used by the navigation system  to produce the vehicle state estimates along with their covariance matrix. 
Note that the state estimate and its covariance matrix will be different at every time step due to noise, changes surrounding environment, and available \ac{GNSS} constellations.  
The navigation system is assumed to be able to provide the state estimate and its covariance in
 \ac{ECEF} (e), intersection \ac{NED} (g), and lane ($\ell$) frame of reference.
Pre-superscripts are used to denote the frame-of-reference for points and vectors.
For example, ${}^e\Boldp$, ${}^g\Boldp$, and ${}^\ell\Boldp$ indicate the position $\Boldp$ represented in the ECEF, intersection, and lane coordinates, respectively.
The frame of reference definitions are reviewed in Appendix \ref{sec:frame}.

The prior state estimate and its error covariance matrix are denoted as 
$$\hat\BoldX_{k}^- = \begin{bmatrix}
	\hat\Boldp_{k}^- & \hat\Boldv_{k}^-
\end{bmatrix}^\top
\mbox{  and }
\BoldC_{\BoldX_{k}}^- = \begin{bmatrix}
	\BoldC_{\Boldp_{k}\Boldp_{k}}^- &\BoldC_{\Boldp_{k}\Boldv_{k}}^-\\
	\BoldC_{\Boldv_{k}\Boldp_{k}}^- &\BoldC_{\Boldv_{k}\Boldv_{k}}^-
\end{bmatrix}.$$

The posterior state estimate and its error covariance matrix are denoted as   
$$\hat{\BoldX}_k^+ =\begin{bmatrix}
	\hat{\Boldp}_k^+ &
	\hat{\Boldv}_k^+
\end{bmatrix}^\top 
\mbox{ and }
\BoldC_{\BoldX_k}^+ = \begin{bmatrix}
	\BoldC_{\Boldp_k\Boldp_k}^+ &\BoldC_{\Boldp_k\Boldv_k}^+\\
	\BoldC_{\Boldv_k\Boldp_k}^+ &\BoldC_{\Boldv_k\Boldv_k}^+
\end{bmatrix}.$$
The  posterior state estimates can be  propagated through time using the state-space model
\begin{align}
\hat{\BoldX}_{k+1}^- &= \Boldphi ~ \hat{\BoldX}_{k}^+   \nonumber\\
\BoldC_{\BoldX_{k+1}}^- &= \Boldphi ~ \BoldC_{\BoldX_{k}}^+ ~ \Boldphi^\top + \BoldQ_d\nonumber
\end{align}
 where $\BoldQ_d$ is the process noise covariance matrix, and 
$\Boldphi = \begin{bmatrix}
\I & \BoldT\\
\0 & \I
\end{bmatrix}$  
where $\BoldT$ is an identity matrix multiplied by the sampling period $\Delta\, t$ such that $\BoldT = \Delta\, t~\I$.

\subsection{Observation $O_k$} \label{sec:Observation}
The observation is the  posterior  vehicle state estimate: $O_k = \hat\BoldX_k^+$.
By the definition of \ac{HMM} observations in \cite{jurafsky_martin}, the observation at time $t_k$ (i.e., $O_k$) depends on the current lane state $q_k$ and is conditionally independent of previous lane states and observations:
\begin{align}	
	P(O_k\,|\,q_0,...,q_k,O_0,...,O_{k-1}) &= P(O_k|q_k). \label{eqn:1}
\end{align}

The vocabulary $\mathcal{V}$ is the set of possible observations at each time step, where each observation $O_k \in \mathcal{V}$.
Throughout this article $\mathcal{V}$ is $\Re^{2}$.
This two dimensional vocabulary corresponds to the roadway surface. 
The height above the roadway surface is not used within this article, but could be directly incorporated. 

The entire sequence of observations is denoted as
\begin{align}
	\mathcal{O} = \{O_k\}_{k=0}^K.		\label{def:observ_seq}
\end{align}
To simplify notation during subsequent derivations,  the following notation:
\begin{align}
	\mathcal{O}_{k_1}^{k_2} &= \{O_{k_1},...,O_{k_2}\} \mbox{ for } k_2>k_1, \label{eqn:OSubSet}
\end{align}
will be used to denote the set of observations between $t_{k_1}$ and $t_{k_2}$.

\subsection{Emission Probability $b_i(O_k)$}\label{subsec:Em}
The symbol $b_i(O_k)$ is the emission probability, which represents the probability of the observation $O_k$ given  that the system is in lane $L_i$:
\begin{align}
	 b_i(O_k) &= P(O_k|q_k=L_i), ~~~0 \leq i \leq N.	 \label{eqn:5}
\end{align}
The emission probability vector $\BoldB_k\in\Re^N$  has elements $[\BoldB_k]_i = b_i(O_k)$.

The emission model used herein is discussed in Section \ref{sec:EmiP}.

\subsection{Transition Probability $a_{ij}^k$}\label{subsec:Tm}
The symbol $a_{ij}^k$ represents the probability of the lane state transitioning from $L_i$  at  $t_k$ to  $L_j$ at time $t_{k+1}$: 
\begin{align}
	a_{ij}^k = P(q_{k+1}=L_j\,|\,q_k=L_i), ~~~0\leq i, j \leq N.\nonumber	 
\end{align}
The lane  state transition matrix $\BoldA_k$ has elements $[\BoldA_k]_{ij} = a_{ij}^k$. 
When the road segment changes, the transition probability must take into account new and disappearing lanes; therefore,  $\BoldA_k$ may not be a square matrix. 
The elements of this transition matrix  must  change with time due to the observations of position and velocity in relation to the high-definition map. 
Therefore, the observation $O_k$ must also be taken into account.
 To achieve this, the transition probability at time $t_k$, conditioned on observation $O_{k}$, is introduced:
\begin{align}
	a_{ij}^k = P(q_{k+1}=L_j|q_{k}=L_i,\,O_{k}), ~~~0\leq i, j \leq N	\label{eqn:4}
\end{align} 
yielding  the data-dependent state transition probability portion  of the  data-dependent  \ac{HMM}.
The  introduced  transition probability has the following properties:
\begin{itemize}
	\item A future lane state only depends on the current lane state and observation:
	\begin{align}
		P(q_{k+1} = L_j\,|\,q_{k} = L_i, q_{k-1},...,q_0,\mathcal{O}_{0}^{k})~~~~~~~~~~~ \nonumber\\
		~~~~~~~= P(q_{k+1} = L_j | q_{k} = L_i,\,O_k).\label{eqn:9}
	\end{align} 
	Conditioning on $q_{k} = L_i$ relates an assumption about which lane contains the vehicle.
	Conditioning on $O_k$ provides specific information related to the position and velocity of the vehicle within the lane that affects the probability of each specific lane change.
	
	\item Each state transition probability must be non-negative:
	\begin{align}
		0 \leq a_{ij}^k \leq 1.\nonumber
	\end{align}
	\item For each $i$, the probability of transitioning from lane state $i$ to some lane state must sum  to 1:
	\begin{align}
		\sum_{j=0}^N a_{ij}^k = 1 ~~~\forall i.\label{eqn:tran_Prop}
	\end{align} 
\end{itemize} 
Section \ref{sec:HMM} follows the definition in eqn. \eqref{eqn:4} through the theory, while reviewing the Viterbi algorithm  to confirm its feasibility. 
Section \ref{sec:TranP} presents the specific  lane transition model used herein.

\subsection{Initial State Probability $\pi^-$}\label{subsec:pi}
The symbol $\pi_i^-$ represents the prior probability, before using any observations, that the system starts in lane $i$: 
\begin{align}
	\pi_i^- = P(q_0=L_i), ~0 \leq i \leq N.\nonumber
\end{align}
The initial state probability vector $\Boldpi^- = [\pi_0^-,...,\pi^-_N].$
Before using any initial measurement, it could be reasonable to assume that the system is equally likely to start in any state.
If that assumption is invoked, then
\begin{align} \label{eqn:pi}
	\pi_i^-  = \frac{1}{N+1}.
\end{align}
Regardless of the assumptions made, the initial state prior probability vector must have positive components that sum to one 
\begin{align}
\| \Boldpi^- \|_1 = \sum_{i=0}^N \pi_i^- = \sum_{i=0}^N P(q_0=L_i)=1.\nonumber
\end{align}

The symbol $\pi_i^+$ represents the  posterior probability of starting in state $i$.
If no observation of the starting position  is available at time $t_k=0$, the  posterior probability is equal to the prior probability (i.e., $\pi_i^+=\pi_i^-$).
If an observation is available at time $t_k=0$, the posterior initial state likelihood\footnote{This is a likelihood vector because it only accounts for the one observation $O_0$, not all possible observations. } denoted as $\bar{\pi}_i^+$ can be computed using the emission probability $b_i(O_0)$ for  observation $O_0$:
\begin{align}
	\bar{\pi}_i^+ &= P(q_0 = L_i,O_0)\nonumber\\
	&= P(O_0|q_0=L_i)P(q_0=L_i)\nonumber\\
	 &= b_i(O_0)~\pi_i^-. \label{eqn:pi_0+}
\end{align}
The likelihood vector $\bar{\pi}_i^+$ can be converted to a probability vector by normalization:
\begin{align}
	\pi_i^+ &= \frac{\bar{\pi}_i^+}{\sum_{i=0}^N \bar{\pi}_i^+}, \label{eqn:pi_norm_batch}
\end{align}
which satisfies $\| \Boldpi^+ \|_1 = 1$.

\section{Viterbi HMM Lane Determination }\label{sec:HMM}
\noindent
This section uses the  data-dependent  \ac{HMM} definitions  presented in  Section \ref{sec:HmmComp} to redefine the Viterbi algorithm  proposed in  \cite{jurafsky_martin,Rabiner}.
 The Viterbi algorithm  is then used to solve the lane determination problem  by finding the most likely lane sequence of hidden states for a given set of observations.
Note that the Viterbi algorithm was originally created for \ac{HMM}; However, the \ac{HMM} proposed herein is a data-dependent \ac{HMM} with a different transition probability definition. Thus, the Viterbi algorithm needs to be redefined to include the data-dependent \ac{HMM}. 
Subsection \ref{subsec:ViterbiAlg} 
redefines the Viterbi algorithm using
the full set of available data $\mathcal{O}$.
Subsection \ref{sect:viterbidisc} discusses the computational requirements of that approach as the number of measurements increase with time.
Subsection \ref{subsec:ViterbiAlgRT} presents real-time Viterbi algorithm solutions to the lane determination problem.

Refer to Table \ref{Table:Viterbi} for symbols used throughout this section.

\begin{table}[b]
	\centering
	\begin{tabular}{|p{1.5cm}||p{6cm}|}
		\hline
		Symbol &Definition \\
		\hline\hline
		$ \mathcal{Q} $ & The lane state sequence of the vehicle \\
		\hline
		$ \mathcal{Q}^* $ & The optimal lane state sequence of the vehicle given the sequence of observations $\mathcal{O}$  \\
		\hline	
		$\delta_k^-(i), \delta_k^+(i) $ & The prior and posterior probability along the most likely sequence of lanes, ending in lane $L_i$ at time $t_k$  \\
		\hline	
		$\psi_k(i) $ &  back trace variable from lane $L_i$ at time $t_k$ \\
		\hline	
		$ q_k^*$ & The optimal lane state at time $t_k$  \\
		\hline	
		$P^* $ &  The joint probability of observations $\mathcal{O}$ and the most probable state sequence $\mathcal{Q}^*$ \\
		\hline	
	\end{tabular}
	\caption{Important symbol definitions used throughout Section \ref{sec:HMM}}
	\label{Table:Viterbi}
\end{table}

\subsection{The Viterbi Algorithm}\label{subsec:ViterbiAlg}
Let $\mathcal{Q} = \{q_k\}_{k=0}^K$ represent a lane state sequence.
Given a time sequence of observations $\mathcal{O}$ as defined in eqn. (\ref{def:observ_seq}) and HMM $\lambda$ as defined in eqn. (\ref{eqn:lambda}), the Viterbi algorithm determines the hidden state sequence $\mathcal{Q}^*=\{q_k^*\}_{k=0}^K$ that is most likely to have produced $\mathcal{O}$ given $\lambda$:
\begin{align}
	\mathcal{Q}^* = \underset{\mathcal{Q}}{\arg\max}~P(\mathcal{Q},\mathcal{O}|\lambda)\label{eqn:Q_Star}.
\end{align}
To solve this maximization, it is convenient to introduce the notation:
 \begin{align}
	\delta^-_k(i) &= \underset{q_0,...,q_{k-1}}{\max} P(q_0,...,q_k=L_i,\mathcal{O}_0^{k-1}|\lambda)	\label{eqn:13}\\
	\delta^+_k(i) &= \underset{q_0,...,q_{k-1}}{\max} P(q_0,...,q_k=L_i,\mathcal{O}_0^k|\lambda),	\label{eqn:20}
\end{align}
where $\mathcal{O}_{k_1}^{k_2}$ is defined in eqn. \eqref{eqn:OSubSet}. 
The variables $\delta^-_k(i)$ and $\delta^+_k(i)$ represents the prior and posterior probabilities along the most likely sequence $(q_0,...,q_{k-1})$, ending in lane $L_i$ at time $t_k$, accounting for the first $k-1$ and $k$ observations respectively.

Using a direct approach, the time complexity of maximizing eqns. (\ref{eqn:13}) and (\ref{eqn:20}) is $O((N+1)^{K})$, where $N$ is the number of physical lanes  and $K$ is the number of observations in the sequence. 
Because the required computational load increases exponentially with K, this approach becomes increasingly infeasible as K increases.
Therefore, a more computationally efficient approach is required.

When the probability along the most probable path $\delta_k^+(i)$ for $0\leq i\leq N$ is known, 
the probability of the most probable path at time $t_{k+1}$ can be computed recursively using a time propagation and measurement update approach to reduce the number of required calculations:
\begin{flalign}
	\delta&^-_{k+1}(j)= \underset{q_0,...,q_{k}}{\max} P(q_0,...,q_{k+1}=L_j,\mathcal{O}_0^k \,|\,\lambda)\label{eqn:23a}\\
	&= \underset{0\leq i \leq N}{\max}\big[
	\underset{q_0,...,q_{k-1}}{\max}P(q_0,...,q_k=L_i,q_{k+1}=L_j,\mathcal{O}_0^k\,|\,\lambda)\big]\label{eqn:16a}\\
	 &= \underset{0\leq i \leq N}{\max}\big[\underset{q_0,...,q_{k-1}}{\max}\big[ P(q_{k+1}=L_j|q_0,...,q_k=L_i,\mathcal{O}_0^k,\lambda)\nonumber\\
	&~~~~~~~~~~~~~~~~~~~~~~~~~~P(q_0,...,q_k=L_i,\mathcal{O}_0^k\,|\,\lambda)\big]\big]\label{eqn:23}\\
	 & = \underset{0\leq i \leq N}{\max}\big[\underset{q_0,...,q_{k-1}}{\max}\big[ P(q_{k+1}=L_j\,|\,q_k=L_i,O_k)\nonumber\\
	&~~~~~~~~~~~~~~~~~~~~~~~~~~P(q_0,...,q_k=L_i,\mathcal{O}_0^k\,|\,\lambda)\big]\big] \label{eqn:24}\\
	 &= \underset{0\leq i \leq N}{\max}[a_{ij}^k~\delta^+_k(i)].\label{eqn:18}
\end{flalign}

Note that eqn. \eqref{eqn:16a} is inferred from eqn. \eqref{eqn:23a} by solving for the maximization of $q_{k}$ first with $\underset{0\leq i \leq N}{\max}$.
The transition probability property described in eqn. \eqref{eqn:9} is then  used to simplify eqn. \eqref{eqn:23} to \eqref{eqn:24}.

The posterior probability $\delta_{k+1}^+(j)$ is calculated as follows
\begin{flalign}	
	&\delta^+_{k+1}(j)=\underset{q_0,...,q_{k}}{\max}P(q_0,...,q_{k+1}=L_j,\mathcal{O}_{0}^{k+1}|\lambda)&&\nonumber\\
	&~~~= \underset{0\leq i \leq N}{\max}[\underset{q_0,...,q_{k-1}}{\max}P(q_0,...,q_{k}=L_i,q_{k+1}=L_j,\mathcal{O}_0^{k+1}|\lambda)].&&\nonumber
\end{flalign}
Note that the main difference between the definition of $\delta_{k+1}^-(j)$ and $\delta_{k+1}^+(j)$ is the inclusion of $O_{k+1}$ in the latter.
Recall from eqn. (\ref{eqn:OSubSet}) that $\mathcal{O}_0^{k+1} = [\mathcal{O}_0^k,O_{k+1}]$, which allows for the following simplification:
\begin{flalign}
	&\delta^+_{k+1}(j) =\nonumber\\
	&~~~\underset{0\leq i \leq N}{\max}[\underset{q_0,...,q_{k-1}}{\max}P(q_0,...,q_{k}=L_i,q_{k+1}=L_j,\mathcal{O}_0^{k},O_{k+1}|\lambda)]\nonumber\\
	&= \underset{0\leq i \leq N}{\max}\big[\underset{q_0,...,q_{k-1}}{\max}\big[ P(O_{k+1}|q_0,...,q_{k}=L_i,q_{k+1}=L_j,\mathcal{O}_0^k,\lambda)\nonumber\\
	&~~~~~~~~~~~~~P(q_0,...,q_{k}=L_i,q_{k+1}=L_j,\mathcal{O}_0^k|\lambda)\big]\big].\nonumber
\end{flalign}
 The observation property described in eqn. \eqref{eqn:1} along with eqn. \eqref{eqn:5} and \eqref{eqn:16a} are  used to get the following simplification:
\begin{flalign}
	&\delta^+_{k+1}(j) = \underset{0\leq i \leq N}{\max}\big[\underset{q_0,...,q_{k-1}}{\max}\big[P(O_{k+1}|q_{k+1}=L_j)\label{eqn:19}\\
	&~~~~~~~~~~~P(q_0,...,q_{k}=L_i,q_{k+1}=L_j,\mathcal{O}_0^k|\lambda)\big]\big]\nonumber\\
	&~~~~= \underset{0\leq i \leq N}{\max}\big[\underset{q_0,...,q_{k-1}}{\max} P(q_0,...,q_{k}=L_i,q_{k+1}=L_j,\mathcal{O}_0^k|\lambda)\big]\nonumber\\
	&~~~~~~~~~~~P(O_{k+1}|q_{k+1}=L_j)\nonumber\\
	&~~~~= \delta^-_{k+1}(j)~b_j(O_{k+1}), \label{eqn:37}
\end{flalign}
The Viterbi algorithm uses the time  propagation  defined by eqn. (\ref{eqn:18}) and measurement update defined by eqn. (\ref{eqn:37}).
This recursive approach reduces the time complexity  from the initial $O((N+1)^K)$  to $O(K(N+1)^2)$, making it  more viable than the direct approach.
Note that adding data dependency to the transition probability does not change the way the Viterbi algorithm is processed  or change its time complexity. 

To keep track of the states that maximize eqn. (\ref{eqn:37}), the back trace variable $\psi_k(i)$ is introduced.
The best sequence of hidden states is generated using the back trace once the final observation has been processed.
The Viterbi Algorithm has the following form:
\begin{itemize}
	\item Initialization: $0\leq i\leq N$ \begin{align}
		\delta^-_0(i) &= \pi_i^-\label{eqn:Init1}\\
		\psi_0(i) &= 0\nonumber\\
		\delta^+_0(i) &= \pi_i^- ~ b_i(O_0)\label{eqn:Init2}
	\end{align}
	\item Time propagation and measurement update: \\
	$~ 0 \leq j \leq N,~ 0\leq k < K-1$ \begin{align}
		\delta^-_{k+1}(j) &= \underset{0\leq i \leq N}{\max}[a_{ij}^k~\delta^+_k(i)]\label{eqn:Induc1}\\
		\psi_{k+1}(j) &= \underset{0\leq i \leq N}{\arg\max}[a_{ij}^k~\delta^+_k(i)]\nonumber\\
		\delta^+_{k+1}(j) &= \delta_{k+1}^-(j)~b_j(O_{k+1})\label{eqn:Induc2} 
	\end{align}
	\item Termination: \begin{align}
		P^* &= \underset{0\leq i \leq N}{\max}[\delta^+_K(i)]\nonumber\\
		q_{K}^* &= \underset{0\leq i \leq N}{\arg\max}[\delta^+_K(i)]\nonumber
	\end{align}
	\item Optimal state sequence backtracking:
	\begin{align}
		q_k^* = \psi_{k+1}(q_{k+1}^*) , ~k= K-1,...,0.\label{eqn:optseq}
	\end{align}
\end{itemize}
The initial state probability $\pi_i^-$ is defined in Section \ref{subsec:pi}.
The emission probability $b_j(O_{k+1})$ is defined in Section \ref{subsec:Em}.
The transition probability $a_{ij}^k$ is defined in Section \ref{subsec:Tm}.
The variable $P^*$ is the joint probability of the observations and most probable state sequence, and $q_K^*$ is the start of the back trace.
The state sequence $\mathcal{Q^*} = \{q_k^*\}_{k=0}^K$ is the optimal state sequence given the sequence of observations $\mathcal{O}$.

\subsection{Viterbi Lane Determination Discussion}\label{sect:viterbidisc}

\begin{figure}[tb]
	\begin{subfigure}[b]{\linewidth}
		\center
		\includegraphics[width=0.6\textwidth,trim={0 0.25cm 0 0.25cm},clip]{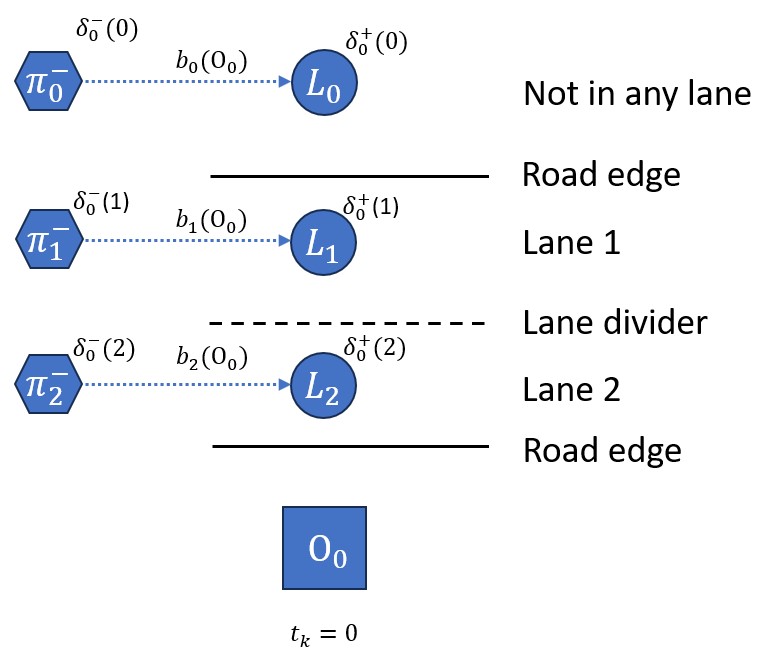}
		\caption{Initialization step}
			\label{fig:Viterbiinit}
	\end{subfigure} 
	
	\vspace{0.25in}
	
	\begin{subfigure}[b]{\linewidth}
		\center
		\includegraphics[width=0.8\textwidth,trim={0 0.25cm 0 0.05cm},clip]{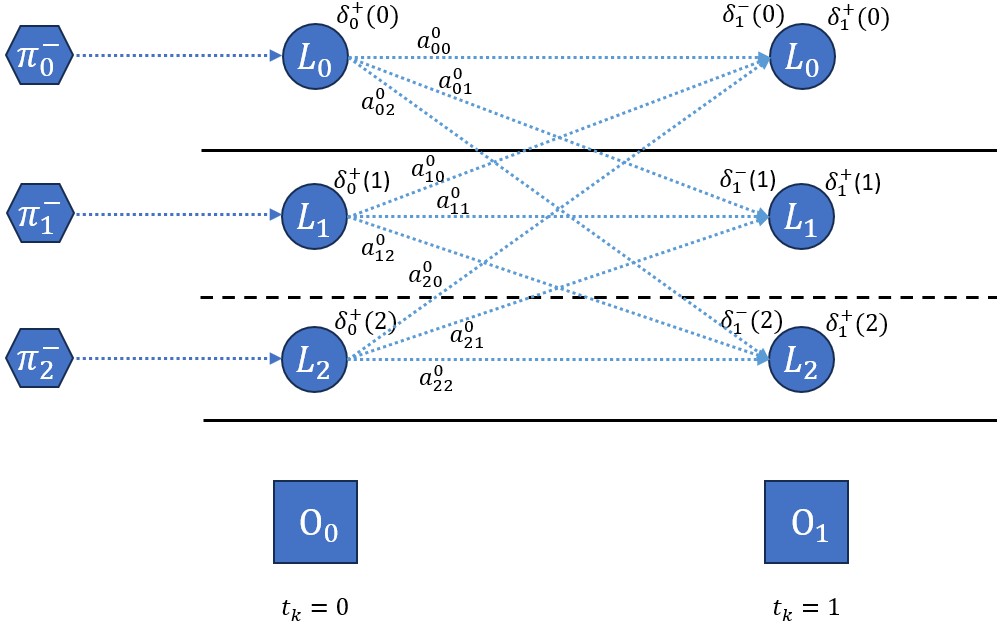}
		\caption{Time propagation and measurement update step}
			\label{fig:Viterbiupdate}
	\end{subfigure}
	
	\vspace{0.25in}
	
	\begin{subfigure}[b]{\linewidth}
		\center
			\includegraphics[width=0.99\textwidth,clip]{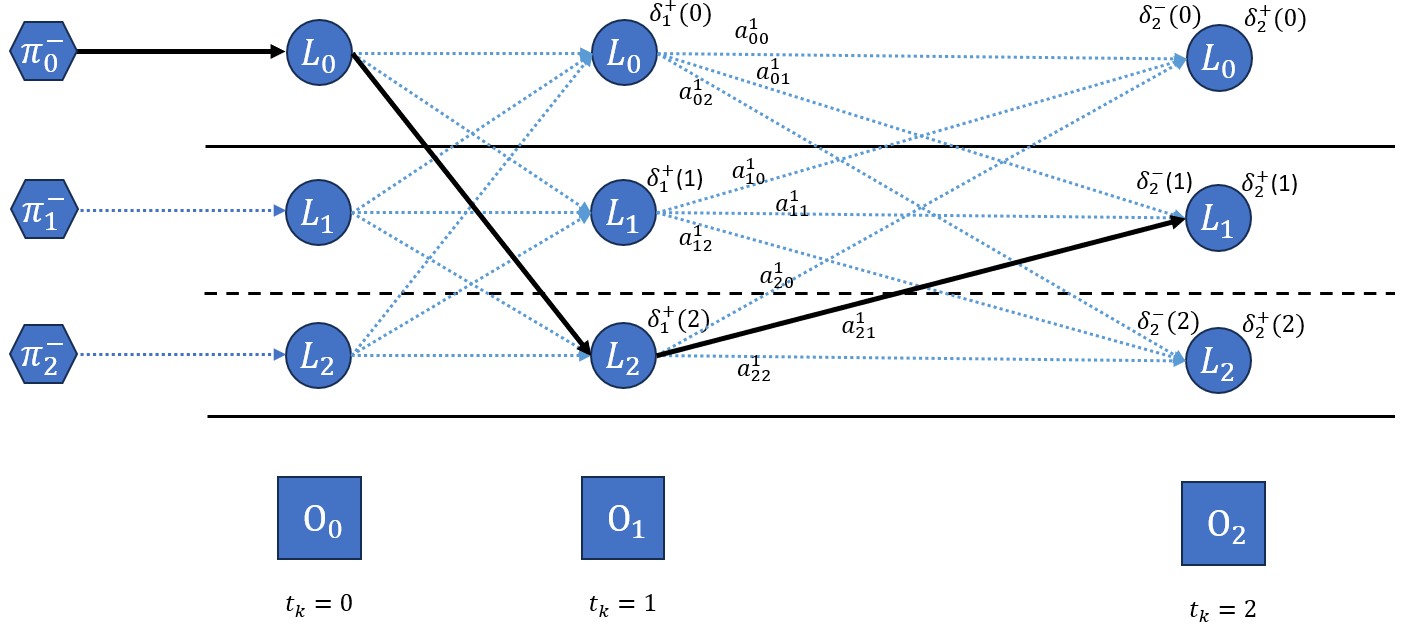}
		\caption{Termination step of the Viterbi algorithm.
			The thick black arrows represent the optimal solution selected by back trace of $\psi_2(q_2^*)$ .}
		\label{fig:ViterbiEnd}
	\end{subfigure}
	\caption{Viterbi lane determination algorithm for a 2 lane road represented as thin black lines. 
		The blue portions show: 
		(a) initialization, 
		(b) time propagation and measurement update, and 
		(c) termination.}
	\label{fig:Viterbi}
\end{figure}

Fig. \ref{fig:Viterbi}  shows a visual representation of the Viterbi lane determination algorithm applied to a two-lane road using three observations with
time increasing from left to right. 
Each circle represents one lane from set $\mathcal{L}_k$, which contains the possible values of the hidden state $q_k$.
The squares represent the observation $O_k$.
The hexagons represent the initial state probability $\pi_i^-$.
The dotted blue arrows represent a multiplication of the variable at the start of the arrow and the variable on the arrow, resulting in the variable at the end of the arrow.

Fig. \ref{fig:Viterbiinit}  corresponds to eqns. \eqref{eqn:Init1}-\eqref{eqn:Init2} in the initialization step.
Fig. \ref{fig:Viterbiupdate} corresponds to eqns. \eqref{eqn:Induc1}-\eqref{eqn:Induc2} in the time propagation and measurement update step. 
Each of these subfigure illustrates how the available variables at the start of the computation (left side) and the observations (bottom)  are used to calculate the results shown on the right.

Fig. \ref{fig:ViterbiEnd} shows the termination step with $\delta^+_2(1)$ maximizing $P^*$. 
The optimal state sequence of the example is $\mathcal{Q^*} = \{q_0^*= L_0,~q_1^*=L_2,~q_2^*=L_1\}$.
The physical interpretation of the example is as follows:
At time $t_k=0$, the vehicle was off the road.
Then at time $t_k=1$, the vehicle merged into lane $L_2$ then did a lane change ending in lane $L_1$ at time $t_k=2$.
Note that $L_0$ does not have a specific off-lane position, so it could be displayed on either side of the road.

Each $\delta_k^+(j)$  is the probability of ending at state $j$ by the most likely path,
based on all observations up to and including time $t_k$. 
For each $j$, its most likely path is distinct. 
Therefore, the vector of probabilities denoted as $\delta_k^+$ is neither a probability vector or a likelihood vector.
The same is true for $\delta_k^-$.
Neither $\|\delta_k^+\|_1$ nor  $\|\delta_k^-\|_1$ should be expected to equal one. 
In fact, as $N$ increases, their 1-norms decrease.
At any time the designer chooses, for example to avoid numeric error due to machine precision, $\delta_k^+$ can be normalized such that its $1$-norm is equal to one (e.g., see eqn. \eqref{eqn:pi_norm_batch}).
The optimal state sequence defined by the maximization in eqn. \eqref{eqn:Q_Star} is unaffected by this normalization.

\subsection{Real-time Viterbi Algorithm}\label{subsec:ViterbiAlgRT}
One approach to implement a real-time algorithm that has fixed computational requirements at each time epoch is to use the Viterbi-based approach of the previous section (as summarized in eqns. \eqref{eqn:Init1}-\eqref{eqn:optseq}) modified as follows.
For $0\le k\le n$, the method of Subsection \ref{subsec:ViterbiAlg} is used with measurements  $\mathcal{O}_{0}^{n}$. 
For $k>n$,  the sliding window of the most recent $(n+1)$ measurements $\mathcal{O}_{k-n}^{k}$ is used. 
Initialization of this approach for each $k>n$ requires determination of the prior probabilities of each lane at time $(k-n)$ (i.e., $\delta_{k-n}^-(j)$).
Two different initialization approaches are as follows.

The first initialization approach uses eqn. \eqref{eqn:pi} to calculate  the initial state probability without any prior information:
\begin{align} 
	\delta_{k-n}^-(j) = \pi_j^- \mbox{ and }
	\delta_{k-n}^+(j) = \pi_j^- ~ b_j(O_{k-n}). \label{eqn:viterbi_init1}
\end{align}

The second approach  calculates a more informed initial state probability for the current window  by propagating the initial state probability  from the previous time window. 
At time step $t_k$, the prior state probability of time step $t_{k-n}$ is defined as
\begin{align}
	\pi_{k-n}^-(j) &\triangleq P(\mathcal{O}_0^{k-n-1}, \, q_{k-n}= L_j) \nonumber \\
	&= \sum_{i=0}^N P(\mathcal{O}_0^{k-n-1}, \, q_{k-n-1}= L_i, \, q_{k-n}= L_j) \nonumber \\
	&=\sum_{i=0}^N P(q_{k-n}= L_j \, | \, \mathcal{O}_0^{k-n-1}, \, q_{k-n-1}= L_i)\nonumber\\
	&~~~~~~~~~~~~~~P(\mathcal{O}_0^{k-n-1}, \, q_{k-n-1}= L_i)  \nonumber 
\end{align}	
\begin{align}
	& = \sum_{i=0}^N P(q_{k-n}= L_j \, | \, q_{k-n-1}= L_i,\, O_{k-n-1})\nonumber\\
	&~~~~~~~~~~P(\mathcal{O}_0^{k-n-1}, \, q_{k-n-1}= L_i)  \label{eqn:25} \\
	&= \sum_{i=0}^N a_{ij}^{k-n-1}~
	 P(\mathcal{O}_0^{k-n-1}, \, q_{k-n-1}= L_i). \label{eqn:RT_pi}
\end{align}
The posterior state likelihood of time step $t_{k-n}$ is defined as \begin{align}
	\bar\pi_{k-n}^+(j) \triangleq P(\mathcal{O}_0^{k-n},q_{k-n}= L_j). \nonumber 
\end{align}
Therefore, $P(\mathcal{O}_0^{k-n-1},q_{k-n-1}= L_i)$ in eqn. \eqref{eqn:RT_pi} is equivalent to $\bar\pi_{k-n-1}^+(i)$.
The fact  that $\mathcal{O}_0^{k-n} = [\mathcal{O}_0^{k-n-1},O_{k-n}]$ allows the following simplification:
	\begin{align}
	&\bar\pi_{k-n}^+(j)= \sum_{i=0}^N P(\mathcal{O}_0^{k-n}, \, q_{k-n-1}= L_i, \, q_{k-n}= L_j) \nonumber \\
	&= \sum_{i=0}^N P(O_{k-n}|\mathcal{O}_0^{k-n-1}, \, q_{k-n-1}= L_i,  \, q_{k-n}= L_j) \nonumber\\
	&~~~~~~~~~~~~~~~~P(\mathcal{O}_0^{k-n-1}, \, q_{k-n-1}= L_i, \,  q_{k-n}= L_j).\nonumber
\end{align}
The properties of eqns. \eqref{eqn:1} and \eqref{eqn:9} allow for the following simplification: 
\begin{align}
	&\bar\pi_{k-n}^+(j)= \sum_{i=0}^N P(O_{k-n}|q_{k-n}= L_j)\nonumber\\
	&~~~~~~~~~~~~P(\mathcal{O}_0^{k-n-1},q_{k-n-1}= L_i, q_{k-n}= L_j) \nonumber\\
	&= \sum_{i=0}^N P(O_{k-n}|q_{k-n}= L_j)~P(\mathcal{O}_0^{k-n-1},q_{k-n-1}= L_i)\nonumber\\
	&~~~~~~~~~~~~P(q_{k-n}= L_j|\mathcal{O}_0^{k-n-1},q_{k-n-1}= L_i)\nonumber\\
	& = P(O_{k-n}|q_{k-n}= L_j)~\sum_{i=0}^N  P(\mathcal{O}_0^{k-n-1},q_{k-n-1}= L_i)~\nonumber\\
	&~~~~~P(q_{k-n}= L_j|q_{k-n-1}= L_i,O_{k-n-1}).\label{eqn:post_pi}
\end{align}
Substituting eqns. \eqref{eqn:5} and \eqref{eqn:25} into \eqref{eqn:post_pi} yields
\begin{align}
	\bar\pi_{k-n}^+(j)= \Bigg[\sum_{i=0}^N a_{ij}^{k-n-1} ~ \pi_{k-n-1}^+(i) \Bigg] b_j(O_{k-n}).\nonumber
\end{align}
The term in square brackets is the time  propagation of the posterior initial likelihood  from the previous time window to the start of the current time window.
The multiplication by $b_j(O_{k-n})$ computes the posterior initial likelihood of each state for the current time window. 
Thus, the second initialization approach initializes \ac{HMM} as \begin{align}
	\delta_{k-n}^-(j) = \pi_{k-n}^-(j) \mbox{ and }
	\delta_{k-n}^+(j) = \bar\pi_{k-n}^+(j). \label{eqn:viterbi_init2}
\end{align}

\section{Lane Model} \label{sec:LaneFrame}
\noindent
To utilize the Viterbi algorithm for lane determination, a lane model is required.
For the model used herein, each road is decomposed longitudinally into road segments, each segment having a constant number of lanes, as described in eqn. \eqref{eqn:laneset_def}. 
Because the navigation system is accurate enough to enable lane determination ( horizontal positional error
$<$ 5m), it is assumed that the vehicle has already determined which road it is on. 

Each lane $L_i$ can be defined by two lane edge curves\footnote{When the roadway model is defined by lane center-lines as in the SAE J2735 standard, lane edge curves can be computed from the lane width and center-lines.}.  
Each curve is defined as a sequence of line segments connecting $M \geq 2$  points:
\begin{equation}
	L_i=\begin{bmatrix}
		{}^g\Boldp_{(u,\,m)}^i\\
		{}^g\Boldp_{(v,\,m)}^i
	\end{bmatrix}, \mbox{ for } m=1,...,M.\label{eqn:LaneStruct}
\end{equation} 
The symbol ${}^g\Boldp_{(u\,,\,m)}^i$ denotes the $i$-th intersection frame point 
in a sequence of connected points defining the left lane edge in the direction of traffic flow of the $i$-th lane.
The points ${}^g\Boldp_{(v\,,\,m)}^i$ are a sequence of connected points defining the right edge of the lane  in the direction of traffic flow of the $i$-th lane. 

\begin{figure}[tb]
	\centering
	\includegraphics[width=0.85\columnwidth]{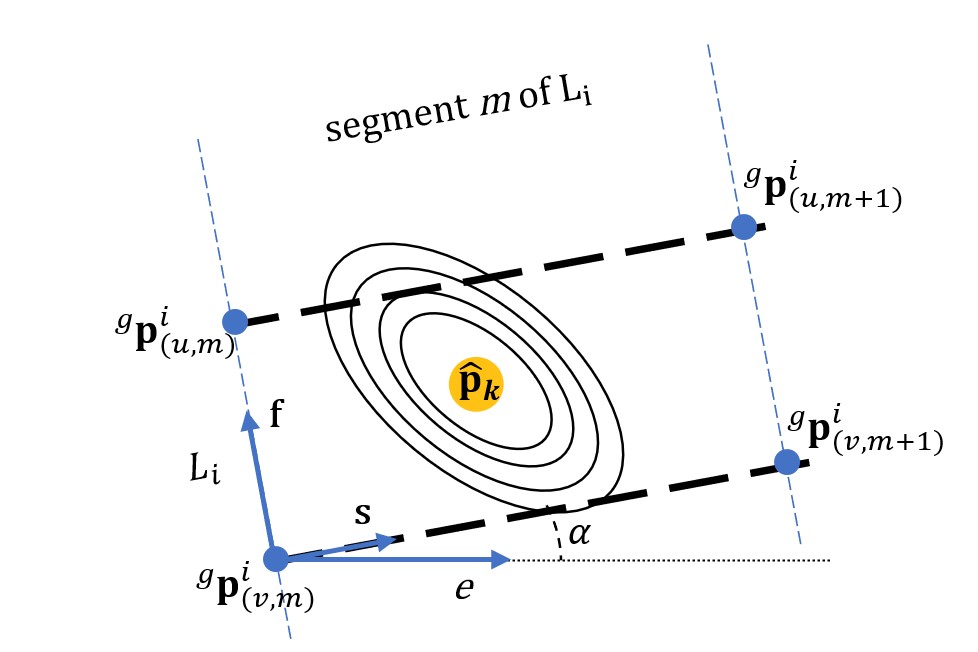}
	\caption{Lane reference frame: road segment m of lane $L_i$ has a reference frame with origin at ${}^g\Boldp_{(v,m)}^i$ and axis $\Boldf$ and $\Bolds$. 
		The symbol $\hat\Boldp_k$ is the estimated vehicle position at time $t_k$. 
		The intersection frame $e$-axis points in the east direction.}
	\label{fig:LaneFrame}
\end{figure}

Fig. \ref{fig:LaneFrame} shows the ellipsoidal probability contours for the position estimate $\hat\Boldp_k$ relative the $m$-th segment of  lane $L_i$. 
Lane frame for  segment $m$ of $L_i$ is defined with origin ${}^g\Boldp_{(v,m)}^i$ and axis directions $\Boldf$ and $\Bolds$.
The vector from ${}^g\Boldp_{(v,\,m)}^i$ to ${}^g\Boldp_{(v,\,m+1)}^i$ defines the $\Bolds$-axis, which is the expected direction of travel for $L_i$.
The $\Boldf$-axis is orthogonal to $\Bolds$-axis and lies in the plane of the road surface, defined positively when pointing towards the left edge of the lane.
The third dimension of the lane frame is marginalized out as it does not affect the vehicle's position within a lane.
In cases where the vertical position of the vehicle is important (e.g., while driving on an overpass), it is assumed that the vertical position is accurate enough to determine the  layer of lanes on which the vehicle is traveling.

The pre-superscript ($\ell$) is used to denote a position vector represented in lane frame.
The pre-subscript ($i$) will used to denote the index of lane $L_i$.
For example, $^\ell_i \hat\Boldp^+_k$ denotes the posterior position estimate $\hat\Boldp^+_k$ represented in lane frame of lane $L_i$.

\section{HMM Probability Models}
\subsection{Lane Emission Probability Model} \label{sec:EmiP}

\noindent
This section discusses the method to compute the emission probability  at time $t_k=kT$ using $O_k$ for the set of lanes $\mathcal{L}_k$ ,  defined in Section \ref{subsec:Em}.

The observation used is  $O_k = {}^\ell_i\hat\Boldp_k^+$  with its error covariance matrix  ${}^\ell_i \BoldC_{\Boldp_k}^+$. 
The symbol  $^\ell_i\Boldp_k$  represents the true, but unknown position of the vehicle.
By its definition in eqn. (\ref{eqn:5}), the emission probability
\begin{align}
	b_i(O_k) &= P(O_k|q_k=L_i) 
	\nonumber\\
	&= \frac{P(q_k=L_i|O_k)~P(O_k)}{P(q_k=L_i)}.
	\label{eqn:EmFrac}
\end{align} 
The subsections that follow will show how to compute $P(q_k=L_i|O_k)$ and $P(q_k=L_i)$.
The factor $P(O_k)$ is unknown, but is identical for all $i \in {0,...,N}$.

Noting the fact that the vehicle has to be either in a lane or off the road, it must be the case that 
\begin{equation}
	\|\BoldB_k\|_1 =	\sum_{i=0}^N b_i(O_k) =  1	\label{eqn:NormT}.
\end{equation}
Therefore, after computing $P(q_k=L_i|O_k)$ and $P(q_k=L_i)$ the quantity 
\begin{align}
	\bar b_i(O_k)
	&= \frac{P(q_k=L_i|O_k)}{P(q_k=L_i)}\label{eqn:EmFracbar}
\end{align}
can be computed and 
the emission probability elements can be normalized
\begin{align}
	b_i(O_k) = \frac{\bar b_i(O_k)}{\sum_{i=0}^N \bar b_i(O_k)} \label{eqn:EmNorm}
\end{align}
to eliminate the unknown  $P(O_k)$.

\subsubsection{Computation of $P(q_k=L_i|O_k)$}

The factor   $P(q_k=L_i|O_k)$ represents the probability that the vehicle is in $L_i$ given the observation $O_k$.
This probability is calculated by integrating the vehicle position probability density over the $i$-th lane.

The vehicle position probability density conditioned on $O_k$ and represented in the $L_i$ frame is 

\begin{align}
	P(^\ell_i\Boldp_k|&^\ell_i\hat\Boldp_k^+)\nonumber\\
	&= \BoldA_{\Boldp_k} \exp\left(-\frac{1}{2}(^\ell_i\Boldp_k-^\ell_i\hat\Boldp_k^+)^\top\, \Big[{}^\ell_i\BoldC_{\Boldp_k}^+\Big]^{-1} (^\ell_i\Boldp_k-^\ell_i\hat\Boldp_k^+)\right)\nonumber
\end{align}
with $\BoldA_{\Boldp_k} = \frac{1}{2\pi ~ |{}^\ell_i\BoldC_{\Boldp_k}^+|^{1/2}}$.
Then, the probability $P(q_k = L_i | O_k)$ is calculated as
	\begin{align}
		P(q_k = L_i | O_k) &= P({}^\ell_i \Boldp_k \in L_i|{}^\ell_i \hat\Boldp_k^+) 
		= \int_{L_i} P({}^\ell_i\Boldp_k|{}^\ell_i\hat\Boldp_k^+)~d\,{}^\ell_i\Boldp_k.\nonumber
	\end{align}
	In lane-frame, $^{\ell}_i\Boldp_k$ has coordinates $[s,f]^\top$
	\begin{align}
		P({}^\ell_i \Boldp_k \in L_i|{}^\ell_i \hat\Boldp_k^+) = \int_{f \in L_i} \left(\int_{S_0}^{S_e} P(\,(s,f) \,|\,^\ell_i\hat\Boldp_k^+\,) ds \right)df \label{eqn:28}
	\end{align}
	where $S_0$ and $S_e$ are the start and end points of the road segment along the $\Bolds$-axis.

	The approximation
	\begin{align}\int_{S_0}^{S_e}P(\,(s,f) \,|\,^\ell_i\hat\Boldp_k^+\,)ds~ \approx \int_{-\infty}^{\infty} P(\,(s,f) \,|\,^\ell_i\hat\Boldp_k^+\,)ds\label{eqn:EmMarg}
	\end{align}
	holds for straight lanes 
	(and lanes where the radius of curvature is much larger than the position uncertainty) 
	since it is assumed that $(S_e-S_0) \gg \sigma_{s}$, where $\sigma_{s}$ represents the position uncertainty along the  $\Bolds$-axis.
	The integral $\int_{-\infty}^\infty P(\,(s,f) \,|\,^\ell_i\hat\Boldp_k^+\,) ~ds$ has the closed form solution
	\begin{align}
		\int_{-\infty}^\infty P(\,(s,f) \,|\,^\ell_i\hat\Boldp_k^+\,)ds=A_f~ \exp\left(-\frac{1}{2}\left(\frac{f-\hat{f}_k^+}{\sigma_{\hat f_k^+}}\right)^2\right),
		\label{eqn:30}
	\end{align}
	where $A_f=\Big(\frac{1}{2\,\pi\,\sigma_{\hat f_k^+}^2}\Big)^{-1/2}$ and the coordinates of ${^\ell_i}\hat\Boldp_k^+$ in lane $L_i$ are defined to be $[\hat{s}_k^+,\hat{f}_k^+]^\top$.  
	Substituting eqn. (\ref{eqn:30}) into eqn. (\ref{eqn:28}) yields 
	\begin{align}
		P({}^\ell_i \Boldp_k \in L_i|{}^\ell_i \hat\Boldp_k^+) &=\int_{0}^{w_i}  A_f \exp\left(-\frac{1}{2}\left(\frac{f-\hat{f}_k^+}{\sigma_{\hat f_k^+}}\right)^2\right) df \label{eqn:22}\\
		&= \Phi_f\left(\frac{w_i-\hat{f}_k}{\sigma_{\hat f_k^+}}\right)-\Phi_f\left(\frac{-\hat{f}_k^+}{\sigma_{\hat f_k^+}}\right),
		\label{eqn:FinalEm}
	\end{align}
	where $w_i$ is the lane width, and $\Phi_f()$ is the \ac{CDF} of a standard normal random variable.

\subsubsection{Computation of $P(q_k=L_i)$}

The factor $P(q_k=L_i)$ represents the  probability of being in lane $L_i$ at time $t_k$ before using observation $O_k$.
Thus, the probability can be calculated using the prior estimate $\hat\Boldp_k^-$ as 
\begin{align}
	P(q_k=L_i) &= P({}^\ell_i \Boldp_k \in L_i|{}^\ell_i\hat\Boldp_k^-)\nonumber\\
	&= \int_{L_i} P({}^\ell_i \Boldp_k | {}^\ell_i \hat\Boldp_k^-) ~ d\,{}^\ell_i\Boldp_k \nonumber\\
	&= \int_{f \in L_i} \left(\int_{S_0}^{S_e} P(\,(s,f) \,|\,^\ell_i\hat\Boldp_k^-\,) ds \right)df\label{eqn:42}.
\end{align}	
The integral in eqn. \eqref{eqn:42} is calculated similarly to eqn. \eqref{eqn:28}. 
Using the approximation of eqn. \eqref{eqn:EmMarg}, the integral $\int_{-\infty}^\infty P(\,(s,f) \,|\,^\ell_i\hat\Boldp_k^-\,) ~ds$ will have the closed form solution 
\begin{align}
	\int_{-\infty}^\infty P(\,(s,f) \,|\,^\ell_i\hat\Boldp_k^-\,)ds=A_f~ \exp\left(-\frac{1}{2}\left(\frac{f-\hat{f}_k^-}{\sigma_{\hat f_k^-}}\right)^2\right).
\end{align}
Then,  $ P({}^\ell_i \Boldp_k \in L_i|{}^\ell_i\hat\Boldp_k^-)$ is calculated as
\begin{align}
	 P({}^\ell_i \Boldp_k \in L_i|{}^\ell_i\hat\Boldp_k^-) &=\int_0^{w_i} A_f~\exp\left(-\frac{1}{2}\left(\frac{f-\hat{f}_k^-}{\sigma_{\hat f_k^-}}\right)^2\right) df 
	 \nonumber\\
	&= \Phi_f\left(\frac{w_i-\hat{f}_k^-}{\sigma_{\hat f_k^-}}\right)-\Phi_f\left(\frac{-\hat{f}_k^-}{\sigma_{\hat f_k^-}}\right).
	\label{eqn:EmDenom}
\end{align}

\begin{figure}[tb]
	\centering
	\includegraphics[scale=0.3]{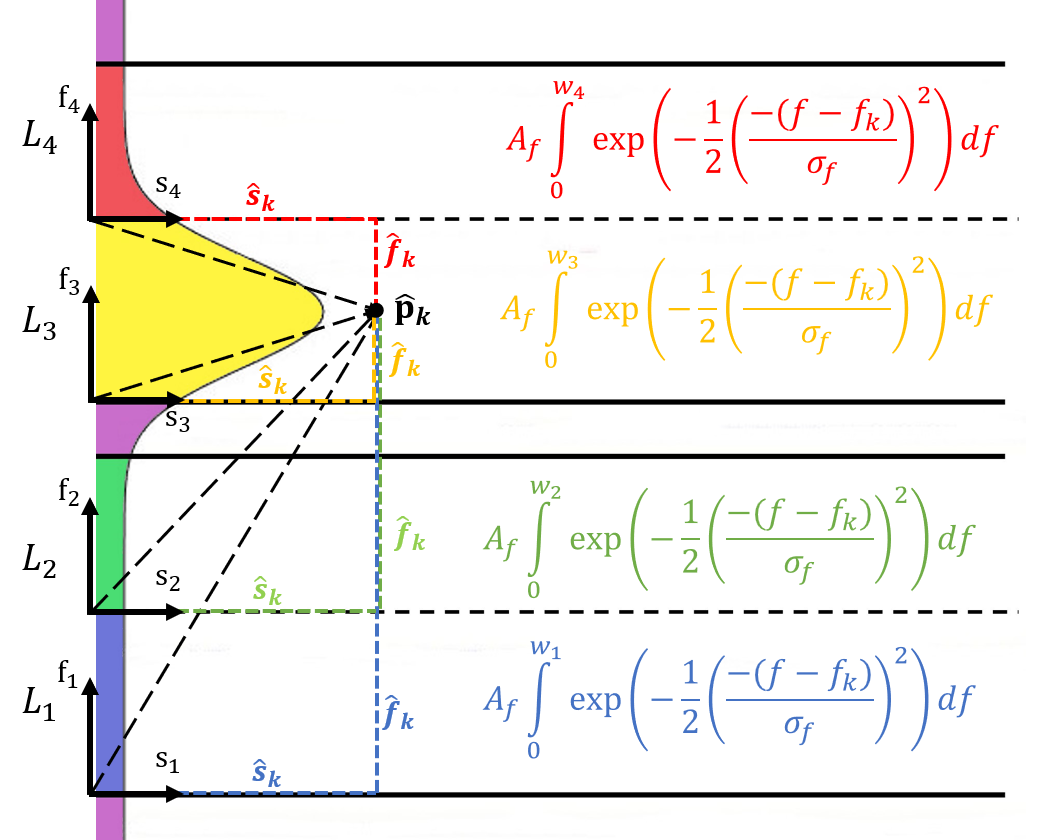}
	\caption{ Illustration of the integral calculations in eqn. \eqref{eqn:FinalEm} or \eqref{eqn:EmDenom} for each lane for position estimate $\hat\Boldp_k$.
		Note that the distribution in the figure was offset to the left from the point $\hat\Boldp_k$ to clearly show each colored area along with the formulas.}
	\label{fig:MarginalDensDist}
\end{figure}

Since $\sum_{i=0}^N P(q_k=L_i|{}^\ell_i\hat\Boldp_k^+) = 1$ when conditioned on either ${}^\ell_i\hat\Boldp_k^-$ or ${}^\ell_i\hat\Boldp_k^+$, the probability of the vehicle being in lane $L_0$, interpreted as not in any lane, can be calculated indirectly 
\begin{align}
	\frac{P(q_k=L_0|O_k)}{P(q_k=L_0)} &= \frac{P(q_k=L_0|{}^\ell_i\hat\Boldp_k^+)}{P(q_k=L_0|{}^\ell_i\hat\Boldp_k^-)}\nonumber\\
	&=  \frac{1-\sum_{i=1}^N P(q_k=L_i|{}^\ell_i\hat\Boldp_k^+)}{1-\sum_{i=1}^N P(q_k=L_i|{}^\ell_i\hat\Boldp_k^-)}.\nonumber
\end{align}
Finally, eqns. \eqref{eqn:EmFracbar} and \eqref{eqn:EmNorm} are calculated to find the normalized emission probability for all states.

Fig. \ref{fig:MarginalDensDist} shows an example of  the integral calculations that are required to calculate emission probabilities   $b_i(O_k)$ for
a road segment with four lanes total, two in each direction, that are separated by a median. 
The black dot denotes the estimated position $\hat\Boldp_k$ with $\hat{f}_k$ and  $\sigma_f^2$  denoting the mean and variance  in the $\Boldf$-direction of a normal distribution.
The lane numbers are indicated along the left edge.
The road edges are marked as solid black lines. 
The lane dividers are marked by dashed black lines.  
The portion of the probability distribution within each lane is indicated by a distinct color:
The blue area accounts for the probability of being in lane $L_1$.
The green area accounts for the probability of being in lane $L_2$.
The yellow area accounts for the probability of being in lane $L_3$.
The red area accounts for the probability of being in lane $L_4$.
The purple area accounts for the probability of being in $L_0$ (i.e., not in any lane).

\subsection{Lane Transition Probability Model} \label{sec:TranP}

\noindent
The transition probability $a_{ij}^k$ defined in Section \ref{subsec:Tm} is calculated at each time $t_k = k\,T$ for every lane pair $L_i$ and $L_j$ with  $i,j \in 0,...,N$.

By the definition of conditional probability, the transition probability defined in eqn. (\ref{eqn:4}) can be rewritten as
\begin{align}
	a_{ij}^k &= \frac{P(q_k=L_i, \, q_{k+1}=L_j,O_k)}{P(q_k=L_i,O_k)}\nonumber\\
	& = \frac{P(q_k=L_i, \, q_{k+1}=L_j|O_k)P(O_k)}{P(q_k=L_i|O_k)P(O_k)}\nonumber\\
	& = \frac{P(q_k=L_i, \, q_{k+1}=L_j|O_k)}{P(q_k=L_i|O_k)}.	\label{eqn:transP}
\end{align}
Eqn. (\ref{eqn:transP}) has two probabilities that need to be calculated. 

\subsubsection{Calculation of $P(q_k = L_i|O_k)$}
\label{sect:aij1}
The quantity $P(q_k= L_i|O_k)$ represents the probability of being in lane $L_i$ given the estimated position ${}^\ell_i\hat \Boldp_k^+$.
The probability can be rewritten as
\begin{align}
	P(q_k = L_i|O_k) = P(\Boldp_k\in L_i|\,{}^\ell_i\hat\Boldp_k^+)
	\label{eqn:a-ij^k_compute}
\end{align}
and calculated using eqn. \eqref{eqn:FinalEm}

\subsubsection{Calculation of $P(q_k=L_i,q_{k+1}=L_j|O_k)$}
\label{sect:aij2}

The quantity $P(q_k=L_i,q_{k+1}=L_j|O_k)$ represents the joint probability of the vehicle being in $L_i$ at time $t_k$ and in $L_j$ at time $t_{k+1}$ given observation $O_k$.
By using the positions $\Boldp_k$, $\Boldp_{k+1}$, the estimated position $\hat\Boldp_k^+$, and the prior predicted position $\hat\Boldp_{k+1}^-$, the joint probability can be computed in lane frame as
\begin{align}
	P(q_k=L_i,q_{k+1}=L_j|&O_k) = P({}^\ell_i\Boldp_k \in L_i, {}^\ell_i\Boldp_{k+1}\in L_j|O_k)\nonumber\\
	&= P({}^\ell_i\Boldp_k \in L_i, {}^\ell_i\Boldp_{k+1}\in L_j|{}^\ell_i\hat\Boldp_k^+,{}^\ell_i\hat\Boldp_{k+1}^-).\nonumber
\end{align}

Within this section let the coordinates of $^{\ell}\Boldp_k$ be defined as $[s,f]^\top$ and those of $^\ell_i\Boldp_{k+1}$ be denoted as $[s_{k+1},f_{k+1}]^\top$.

The joint probability requires integration over lane $L_i$ at time $t_k$ and $L_j$  at time $t_{k+1}$
\begin{align}
	& P({}^\ell_i\Boldp_k \in L_i,{}^\ell_i\Boldp_{k+1} \in L_j|O_k)\nonumber	\\
	&= \int_{L_j}\int_{L_i} P({}^\ell_i\Boldp_k,{}^\ell_i\Boldp_{k+1}|{}^\ell_i\hat\Boldp_k^+,{}^\ell_i\hat\Boldp_{k+1}^-) ~ d\,{}^\ell_i\Boldp_k ~ d\,{}^\ell_i\Boldp_{k+1}  \nonumber\\
	&= \int_{L_j}\int_{L_i} P(s,f, s_{k+1}, f_{k+1}|{}^\ell_i\hat\Boldp_k^+,{}^\ell_i\hat\Boldp_{k+1}^-)~ds ~ df ~ d s_{k+1} ~ d f_{k+1}.	\nonumber
\end{align}
Let the coordinates of the estimated vehicle position at time $t_{k+1}$ be defined as $^\ell_i\hat\Boldp_{k+1}^- = [\hat s_{k+1}, \hat f_{k+1}]^\top$. 
The approximation in eqn. (\ref{eqn:EmMarg}) also holds for the integration over both $s$ and $s_{k+1}$.
Thus, the double integral can be simplified as follows with $dD = ds ~ df ~ ds_{k+1} ~ df_{k+1}$
\begin{align}
	&\int_{L_j}\int_{L_i} P(s,f, s_{k+1}, f_{k+1}|^\ell_i\hat\Boldp_k^+,^\ell_i\hat\Boldp_{k+1}^-)\,dD \nonumber\\
	&=\int_{L_j} \int_{-\infty}^\infty  \int_{L_i} \int_{-\infty}^\infty   P(s,f, s_{k+1}, f_{k+1}|\hat{s}_k, \hat{f}_k,\hat s_{k+1},\hat f_{k+1})~dD\nonumber\\
	&=\int_{L_j} \int_{L_i} P(f, f_{k+1}|\hat{f}_k,\hat f_{k+1})~df~df_{k+1}\nonumber\\
	&=\int_{L_j} \int_{L_i} P(\Boldz|\hat\Boldz)~d\Boldz\nonumber\\
	& = \int_{L_j} \int_{L_i} \BoldA_\Boldz~\exp\left(-(\Boldz-\hat{\Boldz})^\top~{}^\ell_i\BoldC_\BoldZ^{-1}~ (\Boldz-\hat{\Boldz})\right)
	~d\Boldz \label{eqn:twotimejointprobability}
\end{align}  
where 
$\BoldA_\Boldz = \frac{1}{2 \pi |{}^\ell \BoldC_\Boldz|^{\frac{1}{2}}}$, 
$\Boldz = \begin{bmatrix} f, f_{k+1} \end{bmatrix}^\top$,
$\hat\Boldz = \begin{bmatrix} \hat{f}_k, \hat f_{k+1} \end{bmatrix}^\top$, 
$d\Boldz = df \, d{f}_{k+1}$, 
and the error covariance matrix is
\begin{equation}
^\ell_i\BoldC_\Boldz = 		\label{eqn:C_z}
\begin{bmatrix}
	\sigma_{f_{k}}^2&\rho_{f_k f_{k+1}} \sigma_{f_{k}} \sigma_{ f_{k+1}}\\
	\rho_{ f_{k+1} f_k}	\sigma_{ f_{k+1}} \sigma_{f_{k}}& \sigma_{ f_{k+1}}^2
\end{bmatrix}.
\end{equation}

The various symbols in the definition of $^\ell_i\BoldC_\BoldZ$ of eqn. \eqref{eqn:C_z} can be computed from known quantities as follows. 

The state vector estimate and its error covariance matrix 
$\BoldC_{\BoldX_k}$
are defined in Subsection \ref{sec:Observation}.
They are known and can be transformed into any desired reference frame.
The standard deviation $\sigma_{f_{k}}$ can be extracted from the  sub-matrix 
$${}^\ell_i\BoldC_{\Boldp_k\Boldp_k} = \begin{bmatrix}
	\sigma_{s_k}^2 & \rho_{s_k f_k} \sigma_{s_k} \sigma_{f_k}\\
	\rho_{s_k f_k}\sigma_{f_k} \sigma_{s_k} & \sigma_{f_k}^2
\end{bmatrix}.$$ 

The estimated position $\hat\Boldp_k$ and prior prediction of the next position $\hat\Boldp_{k+1}^-$ are related by 
\begin{align}
	\hat\Boldp_{k+1}^- = \hat\Boldp_k + \hat\Boldv_kT \label{eqn:27}.
\end{align}
Therefore, the error covariance matrix between $\hat\Boldp_k$ and $\hat\Boldp_{k+1}^-$ is
\begin{align}
	{}^\ell_i\BoldC_{\Boldp_{k}\Boldp_{k+1}} = {}^\ell_i\BoldC_{\Boldp_k\Boldp_k} + {}^\ell_i\BoldC_{\Boldp_k\Boldv_k}T.\nonumber
\end{align}
The correlation $\rho_{f_k  f_{k+1}}$ can be extracted from  ${}^\ell_i\BoldC_{\Boldp_{k} \Boldp_{k+1}}$.
Defining $\BoldPhi = [\BoldI , ~\BoldI \,T]$,
then \begin{align}
	\BoldC_{\Boldp_{k+1}\Boldp_{k+1}} &= \BoldPhi~\BoldC_{\BoldX_k} \BoldPhi^\top + \BoldQ_{pp}\nonumber\\
	&=\BoldC_{\Boldp_k\Boldp_k} + 2\,T\,\BoldC_{\Boldp_k\Boldv_k}+T^2\BoldC_{\Boldv_k\Boldv_k}+\BoldQ_{pp}\label{eqn:29}
\end{align}
where $\BoldQ_{pp}$ is the covariance matrix for the process noise.
The standard deviation $\sigma_{ f_{k+1}}$ can be extracted from $\BoldC_{\Boldp_{k+1}\Boldp_{k+1}}$.

\begin{figure}[tb]
	\centering
	\includegraphics[width=1\columnwidth]{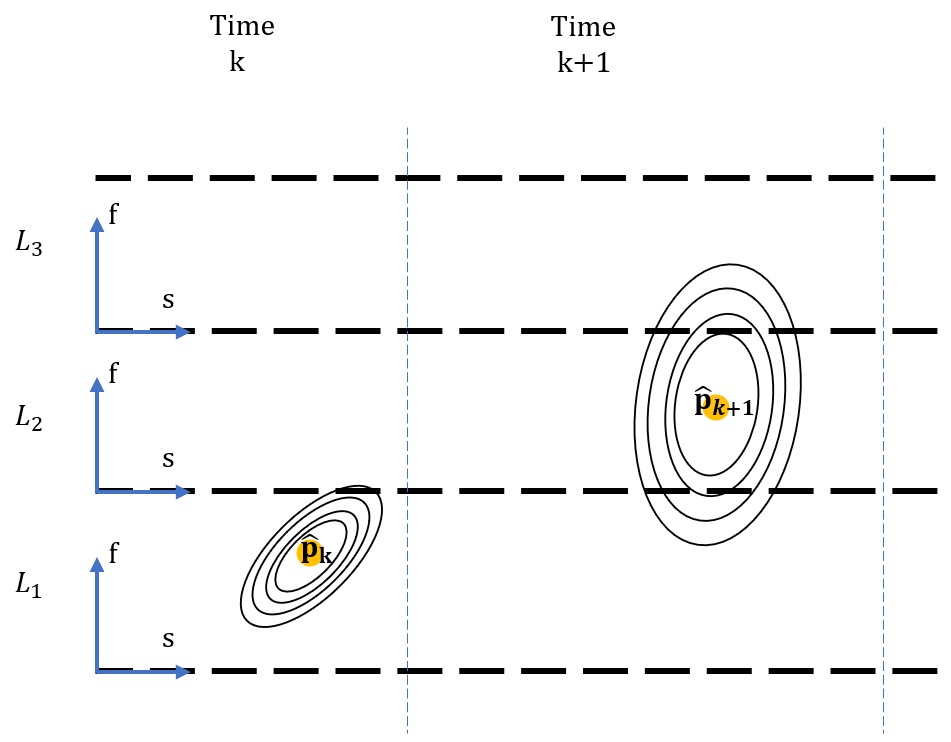}
	\caption{Probability distribution of the estimated vehicle position at times $t_k$ and $t_{k+1}$.}
	\label{fig:twoL}
\end{figure}

Fig. \ref{fig:twoL} depicts  level curves of the probability ellipses to the position at two subsequent times.
The error distribution increases and may rotate during the time advance of eqn. \eqref{eqn:29}.

\begin{figure}[tb]
	\centering
	\includegraphics[width=\columnwidth]{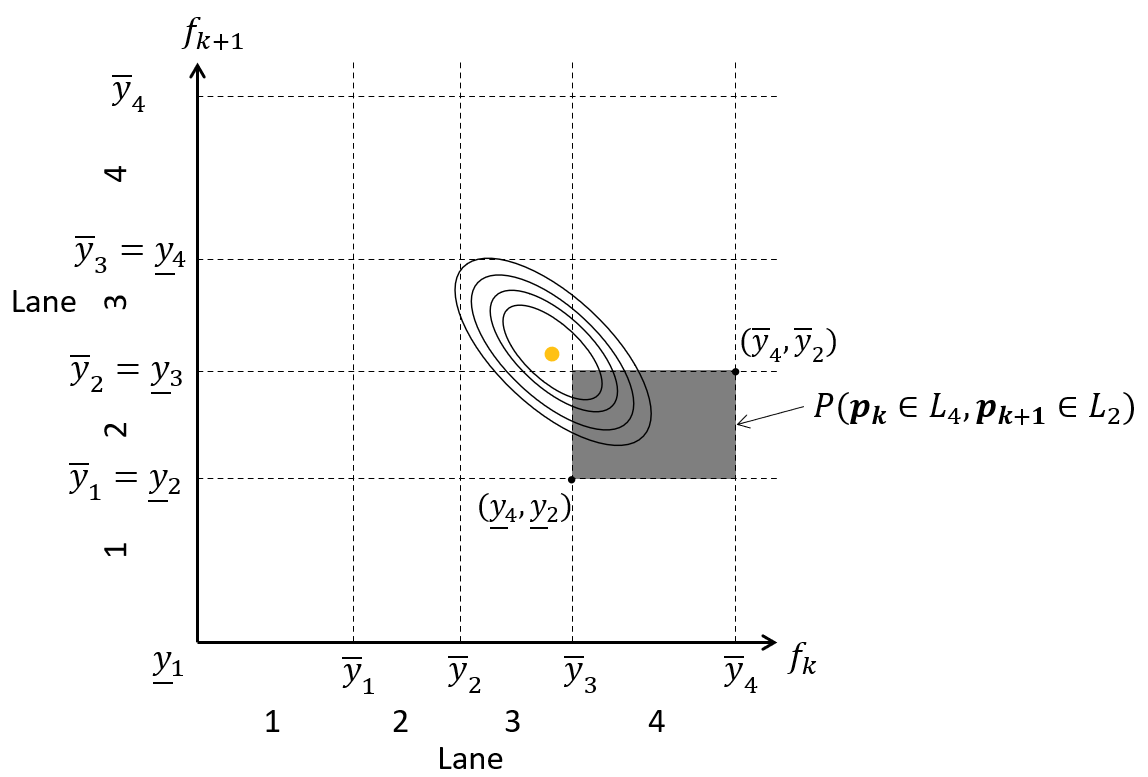}
	\caption{Calculation of the joint probability of being in lane  $L_4$  at time $t_k$ and lane $L_2$ at time $t_{k+1}$, as required for the numerator in Eqn. (\ref{eqn:transP}). 
		The ellipse show the contours of constant probability defined by $\hat{\Boldz}$ and $^\ell_i\BoldC_\Boldz$.}
	\label{fig:JointGraph}
\end{figure}

The joint probability of eqn. \eqref{eqn:twotimejointprobability} can be calculated as
\begin{flalign}
	&P(^\ell_i\Boldp_k \in L_i,^\ell_i\Boldp_{k+1} \in L_j|O_k)\nonumber\\
	&=\int_{0}^{w_j}\int_{0}^{w_i}\BoldA_\Boldz~\exp\left(-(\Boldz-\hat\Boldz)^\top~{}^\ell_i\BoldC_\BoldZ^{-1}~ (\Boldz-\hat\Boldz)\right) d\Boldz \label{eqn:JointP}.
\end{flalign}
Instead of calculating the double integral in eqn. (\ref{eqn:JointP}), the  joint CDF can be used to simplify calculations for each lane pair $L_i$ and $L_j$.
Recall from Section \ref{sec:LaneFrame}, each lane is defined by a sequence of line segments defining the left and right edge of the lane.
Using eqn. \eqref{eqn:LaneStruct}, let $\underline y_i$ and $\bar y_i$ be the $f$ coordinate of ${}^\ell_1\Boldp_{v,m}^i$ and ${}^\ell_1\Boldp_{u,m}^i$ respectively, defined in lane frame of $L_1$.
Then, the joint \ac{CDF} is defined as 
\begin{align}
	& P({}^\ell_i \Boldp_k \in L_i,{}^\ell_i \Boldp_{k+1} \in L_j|O_k) = \Phi(\bar y_i , \bar y_j , \hat\Boldz , {}^\ell_i\BoldC_{\Boldz})  \label{eqn:39}\\
	&
	-\Phi(\bar y_i , \underline y_j , \hat\Boldz , {}^\ell_i\BoldC_{\Boldz})
	-\Phi(\underline y_i , \bar y_j , \hat\Boldz , {}^\ell_i\BoldC_{\Boldz}) 
	+ \Phi(\underline y_i , \underline y_j , \hat\Boldz , {}^\ell_i\BoldC_{\Boldz}).\nonumber
\end{align}
The \ac{CDF} calculation of eqn. \eqref{eqn:39} is illustrated in Fig. \ref{fig:JointGraph}.
Note that in the figure, all lanes share a lane edge with no median in between.
Thus, the following equality holds $\underline y_i = \bar y_{i-1}$ for $i= 2,3,4$.

Each lane at time $t_k$ corresponds to a column of rectangles.
Each lane at time $t_{k+1}$ corresponds to a row of rectangles. 
The height and width of each rectangle match the lane width. 
The shaded region corresponds to being in lane  $L_4$  at time $t_k$ and lane $L_2$ at time $t_{k+1}$.
Performing the integration of eqn. \eqref{eqn:JointP} or the \ac{CDF} calculation of eqn. \eqref{eqn:39} for this shaded region provides the probability of transitioning from lane   $L_4$  at time $t_k$ to lane $L_2$ at time $t_{k+1}$.

\subsubsection{Calculations of $a_{ij}^k$ Containing Lane $L_0$}

Sections \ref{sect:aij1} and \ref{sect:aij2} computed $a_{ij}^k$ for lanes $i,\,j=1,\ldots,N$  directly. 
The transition probabilities related to $L_0$ must be calculated indirectly.

\subsubsection{The probability of transitioning to $L_0$}
The probability of transitioning to $L_0$ represents the probability of transitioning to an off road position and is calculated using eqn. \eqref{eqn:transP}:
\begin{align}
	a_{i \, 0}^k = \frac{P(q_k=L_i, \, q_{k+1}=L_0|O_k)}{P(q_k=L_i|O_k)} \nonumber.
\end{align}
To calculate the numerator, the following property is used
\begin{align}
	\sum_{j=0}^N P(q_k=L_i, \, q_{k+1}=L_j|O_k) = P(q_{k}=L_i |O_k) \nonumber.
\end{align}
The numerator is calculated as
\begin{align}
	P(q_k = L_i, &q_{k+1} = L_0|O_k) &	\nonumber\\
	&= P(q_k=L_i | O_k) - \sum_{j=1}^N P(q_k=L_i, \, q_{k+1}=L_j|O_k), 	\nonumber
\end{align}
where 
$P(q_{k}=L_i | O_k)$  is known from eqn. \eqref{eqn:a-ij^k_compute}
and
$P(q_k=L_i, \, q_{k+1}=L_j|O_k)$  is known from eqn. \eqref{eqn:39} for $j=1,\ldots,N$.
The denominator is calculated using eqn. \eqref{eqn:FinalEm} as discussed in Section \ref{sec:EmiP}.

\subsubsection{The probability of transitioning from $L_0$}
The probability of transitioning from $L_0$ represents the probability of transitioning from an off-road position to a lane on the road and is calculated using eqn. \eqref{eqn:transP} as:
\begin{align}
	a_{0 \, j}^k = \frac{P(q_k=L_0, \, q_{k+1}=L_j|O_k)}{P(q_k=L_0|O_k)} \nonumber.
\end{align}
The denominator is calculated using eqn. \eqref{eqn:FinalEm} as discussed in Section \ref{sec:EmiP}.
To calculate the numerator, the following property is used
\begin{align}
	\sum_{i=0}^N P(q_k=L_i, \, q_{k+1}=L_j|O_k) = P(q_{k+1}=L_j |O_k) \nonumber.
\end{align}
Then the numerator is calculated as
\begin{align}
	P(q_k = L_0,&\,q_{k+1} = L_j|O_k)  \nonumber \\
	&= P(q_{k+1}=L_j | O_k)- \sum_{i=1}^N P(q_k=L_i, \, q_{k+1}=L_j|O_k),		\nonumber
\end{align}
where 
$P(q_{k+1}=L_j | O_k)$  is known from eqn. \eqref{eqn:a-ij^k_compute}
and
$P(q_k=L_i, \, q_{k+1}=L_j|O_k)$  is known from eqn. \eqref{eqn:39} for $j=1,\ldots,N$.

\subsubsection{The probability of transitioning from $L_0$ to $L_0$}
The probability of transitioning from $L_0$ to $L_0$ represents the probability of staying off the road and is calculated using the property in eqn. \eqref{eqn:tran_Prop} such that
\begin{align}
	a_{0 \, 0}^k = 1 - \sum_{j=1}^N a_{0 \, j}^k\nonumber.
\end{align}

\section{Non-Parallel Lanes Determination and Road Segment Selection}\label{sec:NP_Lanes} 
\noindent
The previous sections have only considered the case where all lanes are parallel. 
This section considers the situation where some lanes are not parallel to the others such as road layouts with merging lanes or highway exits.

As discussed in Section \ref{sec:LaneFrame}, the emission and transition probabilities are calculated in each lane's frame.
When the lanes are parallel, the transformations between each lane frame is only a translation of the origin.
In that special case,   eqns. \eqref{eqn:FinalEm} \eqref{eqn:EmDenom}, and \eqref{eqn:JointP} can be calculated in the same lane frame.

However, in road layouts where the lanes are not parallel, the emission and transition probability models need to be modified to account for different lane frame orientations and translations. 

The emission probability for a non-parallel lane is calculated by eqn. \eqref{eqn:EmFracbar}. 
Since the lane frames are not parallel,  the probability distribution computation of $P(q_k=L_i|O_k)$ and $P(q_k=L_i)$ will be different in every non-parallel lane frame since the covariance is rotated to that specific frame.

Similarly, the transition probability is modified to account for lane frame orientation changes.
Recall from Section \ref{sec:TranP} that $a_{ij}^k$ has the joint probability $P(q_k=L_i,q_{k+1}=L_j|O_k)$ which needs to be calculated.
In order for the joint probability defined in eqn. (\ref{eqn:JointP}) to work, the covariance ${}^\ell \BoldC_\Boldz$ needs to be rotated to the proper frame.
The rotation matrix ${}_g^\ell \BoldR_{im}$ is used to rotate from g-frame to $\ell$-frame of lane segment m of lane $L_i$ , defined in  Appendix \ref{sec:frame}. 
The covariance ${}^\ell \BoldC_\Boldz$
is marginalized from $^\ell_i \BoldC_{\Boldp_k\Boldp_k}$, $^\ell_i \BoldC_{\Boldp_k\Boldp_{k+1}}$, and $^\ell_j \BoldC_{\Boldp_{k+1}\Boldp_{k+1}}$ as described in eqn. \eqref{eqn:C_z}.
Note that $^\ell_j \BoldC_{\Boldp_{k+1}\Boldp_{k+1}}$ is rotated to $L_j$ while the rest of the covariances are rotated to $L_i$ using eqn. \eqref{eqn:CovMatrix} defined in Appendix \ref{sec:frame}.
Similarly, $\hat\Boldp_k$ and $\hat\Boldp_{k+1}^-$ need to be rotated to $L_i$ and $L_j$ respectively using eqn. \eqref{eqn:PointRotate} defined in Appendix \ref{sec:frame}.
Then, the new $\hat\Boldz = [\hat f_k, \hat f_{k+1}]$ and $^\ell \BoldC_\Boldz$ are used to calculate $a_{ij}^k$ using eqn. \eqref{eqn:JointP}.
If there was a need to calculate all variables in the same frame, the mentioned modifications can also be used in parallel lanes without affecting the emission and transition probabilities.

\label{sec:LD_model}
To calculate the HMM probabilities, the lane frame model defined in Section \ref{sec:LaneFrame} uses $m$ to index the road segments.
Because the  road segment index of the vehicle at each time is not known, the index $m$ has to be calculated.

For each road segment $1\le m < M$, let 
$^\ell\hat\Boldp_k^m = \left( s_k^m, f_k^m\right)$ 
denote the lane coordinates of $\hat\Boldp_k$ for the first lane of the segment. 
The vehicle is on segment  $m$ if and only if $s_k^m\ge 0$ and $s_k^{m+1}<0$.

\section{Experiment}\label{sec:Analysis}
\noindent
The goal of this section is to demonstrate and analyze the experimental performance of the \ac{HMM} and Viterbi methods presented herein in comparison with other methods in the literature.

\subsection{Data Collection Setup and Discussion} \label{sec:Experiment}

To support the experimental analysis GNSS vehicle position and velocity estimates are acquired while a driver performs various lane changes and vehicle maneuvers (e.g., U-turns).
Two datasets were collected on Columbia Avenue in Riverside, California between the intersections with  Iowa Avenue and Northgate Street.
The driving speed varied from 0 through 45 mph, with typical speeds between 35-45 mph. 
The driving for data acquisition  of a dataset  lasted for approximately 32 minutes and included multiple lane changes and U-turns.
Throughout the test, the satellite visbility was good (i.e., under open sky conditions).

Two u-blox ZED-F9P dual-frequency receivers were mounted on the vehicle for the experiment.
Each was connected to the same u-blox antenna via a signal splitter.
The first receiver performs \ac{DGNSS} using GPS, GALILEO, BeiDou satellites and SBAS corrections.
It provides the state estimates that comprise $O_k$.
The second receiver is communicating with a local base station. 
It uses GPS, GALILEO, and BeiDou satellites to generate its native multi-GNSS, dual frequency, \ac{RTK}-fixed solution, which uses carrier phase measurements to achieve centimeter-level accuracy \cite{teunissen2017springer}.  

Given this high-level of position accuracy the RTK data was used as the ground truth for the experiment, from which the `correct' lane sequence could be determined directly. 
This ground truth lane sequence was used to evaluate correctness of the lane sequences determined by each method based on the data from the first receiver.
The RTK data was not otherwise used in any of the results.

To extract $\BoldC_{\Boldp_k}$ and $\BoldC_{\Boldv_k}$ in g-frame, the receiver binary output was  parsed to decode the UBX-NAV-COV protocol, which contains the required covariances.
The covariance $\BoldC_{\Boldp_k \Boldv_k}$ is not part of the UBX-NAV-COV protocol.
Because position is computed from pseudorange and velocity is computed from Doppler,  herein those measurements are assumed to have independent measurement noise, so that $\BoldC_{\Boldp_k \Boldv_k}$ was assumed to be a zero matrix.

\subsection{Experiment Datasets}

 \label{sec:Datasets}
The different lane determination models will be compared using two experiments with two datasets denoted DS 1 and DS 2.
DS 1 contains 2045 points while DS 2 contains 1862 points.
The data sample rate is one Hertz.

Let ${}^g\Boldp_k = \begin{bmatrix}
	n_k\\
	e_k
\end{bmatrix}$ and ${}^g\hat\Boldp_k = \begin{bmatrix}
\hat n_k\\
\hat e_k
\end{bmatrix}$,
the horizontal error at time $t_k$ will be calculated as $e_{h_k} = \sqrt{e_{n_k}^2+e_{e_k}^2}$	where the north and east errors at time $t_k$ are $e_{n_k} = (\hat n_k- n_k)$ and $e_{e_k} =(\hat e_k - e_k)$.
The mean horizontal error will be calculated as
\begin{align}
	\text{mean error} = \frac{1}{N}\sum_{k=1}^N e_{h_k}\nonumber.
\end{align}
Dataset DS1 has a minimum and maximum horizontal error of 0.03 m and 1.26 m respectively and a mean horizontal error of 0.58 m.
Dataset DS2 has a minimum and maximum horizontal error of 0.08 m and 2.55 m respectively and a mean horizontal error of 0.53 m.

DS 1 contains more instances where the vehicle was not in any lane such as when performing a U-turn or entering an intersection. 
Since the methods that already exist in the literature do not account for off-road positions directly during the emission and transition probability calculations the comparison will use two experiments:
\begin{description}
	\item[Experiment 1] uses the two collected datasets. 
	Each dataset includes a complete drive cycle with lane changes and off-road positions.
	DS 1 contained 192 lane changes, while DS 2 contained 159 lane changes.
	
	\item[Experiment 2] uses a modified version of the two datasets that excludes off-road positions:
	199 points are excluded
	for DS 1, while  61 points are excluded for DS 2.
\end{description}  
\subsection{Tested Methods}
 Five methods will be tested using the experiments and datasets described in Section \ref{sec:Datasets}:

\begin{description}
	\item [Proposed Method] is the Viterbi approach of Section \ref{sec:HMM} using a \ac{HMM} with the  emission and transition probability models defined in Sections  \ref{sec:EmiP} and \ref{sec:TranP}.
	
	\item [Method 2] is the Viterbi approach of Section \ref{sec:HMM} using a HMM with the emission probability model proposed in \cite{Hannson2021} and transition probability model defined in Section \ref{sec:TranP}.
	
	\item [Method 3] is the Viterbi approach of Section \ref{sec:HMM} using a \ac{HMM} with the  emission probability model defined in Section \ref{sec:EmiP} and the transition probability model of \cite{Hannson2021}.

	\item [Method 4] is the Viterbi approach of Section \ref{sec:HMM} using a \ac{HMM} with the   emission and transition probability models proposed in \cite{Hannson2021}.

	\item [Method 5] uses the distance from the estimated vehicle position to the lane center-line as an error function to implement the approach defined in \cite{Atia2017,Fu2019}.
	Let ${}^c \Boldp_k^j$ represent the projection of the estimated position $\hat\Boldp_k$ onto the lane center-line of $L_j$.
	The error function used the observations $\mathcal{O}_{k-d+1}^{k}$ at a time $t_k$, where $d$ is the number of observations used.
	The error function is
	\begin{align}
		err(L_j) = \frac{1}{d} \sum_{i=1}^d ||
		 {}^\ell_j
		\hat\Boldp_{k-d+1} - {}^c\Boldp^j_{k-d+1}||_2.
	\end{align}	
	The sequence of lanes $\mathcal{Q}_{k-d+1}^{k}$ is set to the one lane $L_j$ which minimizes the error function such that 

	\begin{align}
		\mathcal{Q}_{k-d+1}^{k} = \underset{0 \leq j \leq N}{\arg\min} 
		~err(L_j).
	\end{align}
\end{description}

The results in \cite{Hannson2021}  were achieved using  multiple sensors.
Therefore, the emission model in \cite{Hannson2021} included factors for each sensor modality. 
That dataset is no longer available. 
The dataset used herein only includes \ac{GNSS} data. 
Therefore, the emission model used herein for  Methods 2 and 4  only uses the GNSS observation factor of the emission model in \cite{Hannson2021}.

Methods 3 and 4 both use the transition probability of \cite{Hannson2021} which does not account for the case of the vehicle not being in a physical lane. 
When the vehicle is not in a lane (e.g., during a U-turn), these \ac{HMM} methods may break  during the Viterbi algorithm calculations because $\delta_k^+(i) = 0 ~\forall i$.  
Herein, we implement a resetting approach referenced by \cite{Hannson2021,Newson} to accommodate these HMM breaks  in Methods 2, 3, and 4. 
When a break is detected, the data is split into separate trips, and the \ac{HMM} is reinitialized by recalculating eqns. (\ref{eqn:Init1}) and (\ref{eqn:Init2}).
After the data split, the Viterbi algorithm is run separately on each trip.
The trip paths are then concatenated to generate the complete path.
The complete path is then compared to the ground truth to evaluate the accuracy of a given model.
At every break, the state at that time step is unknown, and therefore counted as  incorrect when evaluating accuracy.

\subsection{Viterbi Algorithm Processing}
For each experiment, the Viterbi algorithm summarized in eqns. (\ref{eqn:Init1}-\ref{eqn:optseq}) processes observations using three different approaches.
\begin{description}
	\item[Approach 1] processes the entire sequence of observations $\mathcal{O}$ once at the end of the trip.
	\item[Approach 2] processes the most recent five observations $\mathcal{O}_{k-4}^{k}$ at each $t_k$ with the initialization at each $t_{k-4}$ defined by eqn. \eqref{eqn:viterbi_init1}.
	\item[Approach 3] processes the most recent five observations $\mathcal{O}_{k-4}^{k}$ at each $t_k$  with the initialization at each $t_{k-4}$ defined by eqn. \eqref{eqn:viterbi_init2}.
\end{description}

\subsection{Analysis of the Results}

Table \ref{Table:Experm1Result} presents the results from Experiment 1 using the first four methods and both data sets with approximate 95\% confidence level.
The confidence interval was calculated using the method described in Section 8.1 of \cite{moore2009introtostats}.

The results for the three approaches are presented in separate sub-tables. 
\begin{itemize}
	\item  Proposed Method  achieved the highest accuracy in all cases. 
	
	\item  Proposed Method  had no breaks. 
	Each of the other methods had several breaks. 
	There are more breaks for DS 1 because it  contains more instances were the vehicle was not in any lane.
	The number of \ac{HMM} breaks highlights the importance of accounting for off-road locations in lane determination as all methods that do not account for off-road positions in both the emission and transition probabilities suffered from \ac{HMM} breaks.
		
	\item Method 3 used the emission model herein that accounts for off-road positions, while its transition model does not.
	Such discrepancy affects how the Viterbi algorithm is propagated through time as the maximization in eqn. \eqref{eqn:Induc1} might not be the optimal solution due to the model's inability to calculate transitions from an off road position, thus lowering the performance.
	
	\item Method 4 uses the emission and transition models of \cite{Hannson2021}.
	Changing only the transition model of \cite{Hannson2021} to that proposed herein (i.e., Method 2) yielded an average accuracy increase of  $1.8\%$ and $2.8\%$  for Datasets 1 and 2.

\end{itemize}

Table \ref{Table:Experm2Result} presents the results from Experiment 2 (excluding off-road positions)
using the first four methods and both datasets.
The results for each approach are presented in a separate sub-tables.
\begin{itemize}
	\item  In Approaches 1 and 3,  Proposed Method, Method 2, and Method 4  achieve similar levels of accuracy with Method 2 achieving the highest accuracy within margin of error of the Proposed Method. 
	\item In Approach 2, Proposed Method  achieves the best accuracy by  approximately  10\%.
	\item  Proposed Method  experiences no breaks, whereas all of the other Methods  still  experience breaks even though all off-road positions have been excluded. 
	This suggests that \ac{HMM} breaks are still possible based on the emission and transition probability model combinations used in Methods 2-4 and are not only caused by off-road positions. 
	\item Method 4 uses the emission and transition models of \cite{Hannson2021}.
	Changing only the transition model of \cite{Hannson2021} to that proposed herein (i.e., Method 2) yielded an average accuracy increase of $1.9\%$ for both Datasets 1 and 2.
\end{itemize}

\begin{table}[t]
	\centering
	\resizebox{\columnwidth}{!}{
	\begin{tabular}{|c | c c | c c|}
		\multicolumn{5}{c}{Approach 1}\\
		\hline
		 & Accuracy &  Number of Breaks & Accuracy & Number of Breaks\\
		\hline
		\T\B Proposed Method & $\mathbf{95.6\% \pm 0.9\%}$ & 0 &  $\mathbf{95.1\% \pm 1.0\%}$ & 0 \\
		\hline
		\T\B Method 2 &  $88.4\% \pm 1.4\%$ & 144 & $93.9\% \pm 1.1\%$ & 20  \\
		\hline
		\T\B Method 3 & $61.2\% \pm 2.1\%$ & 172 & $74.6\% \pm 2.0\%$ & 42 \\
		\hline
		\T\B Method 4 & $84.6\% \pm 1.6\%$ & 170 & $89.6\% \pm 1.4\%$ & 43 \\
		\hline
		\multicolumn{1}{c}{}\\
		\multicolumn{5}{c}{Approach 2}\\
		\hline
		 & Accuracy &  Number of Breaks & Accuracy &  Number of Breaks\\
		\hline
		\T\B Proposed Method &  $\mathbf{96.0\% \pm 0.9\%}$ & 0 &  $\mathbf{95.0\% \pm 1.0\%}$ & 0  \\
		\hline
		\T\B Method 2 & $88.3\% \pm 1.4\%$ & 139 & $93.9\% \pm 1.1\%$ & 16 \\
		\hline
		\T\B Method 3 & $71.2\% \pm 2.0\%$ & 172 & $77.9\% \pm 1.9\%$ & 20 \\
		\hline
		\T\B Method 4 & $86.9\% \pm 1.5\%$ & 175 & $92.2\% \pm 1.2\%$ & 41\\
		\hline
		\multicolumn{1}{c}{}\\
		\multicolumn{5}{c}{Approach 3}\\
		\hline
		 & Accuracy & Number of Breaks & Accuracy &  Number of Breaks\\
		\hline
		\T\B Proposed Method & $\mathbf{96.3\% \pm 0.8\%}$ & 0 & $\mathbf{95.1\% \pm 1.0\%}$ & 0  \\
		\hline
		\T\B Method 2 & $88.3\% \pm 1.4\%$ & 140 & $93.9\% \pm 1.1\%$ & 16 \\
		\hline
		\T\B Method 3 & $69.2\% \pm 2.0\%$ & 297 & $70.7\% \pm 2.1\%$ & 136\\
		\hline
		\T\B Method 4 & $86.2\% \pm 1.5\%$ & 301 & $91.5\% \pm 1.3\%$ & 146\\
		\hline
	\end{tabular}}
	\caption{Experiment 1 results.
		Columns 2 and 3 present the results using DS 1, while columns 4 and 5 present the results using DS 2.
	}
	\label{Table:Experm1Result}
\end{table}

\begin{table}[t]
	\centering
	\resizebox{\columnwidth}{!}{
	\begin{tabular}{|c | c c | c c|}
		\multicolumn{5}{c}{Approach 1}\\
		\hline
		& Accuracy & Number of Breaks & Accuracy &  Number of Breaks\\
		\hline
		\T\B Proposed Method & $96.9\% \pm 0.8\%$ & 0 &  $95.8\% \pm 0.9\%$ & 0 \\
		\hline
		\T\B Method 2 &  $\mathbf{97.6\% \pm 0.7\%}$ & 8 & $\mathbf{96.4\% \pm \0.9\%}$ & 3  \\
		\hline
		\T\B Method 3 & $65.3\% \pm 2.2\%$ & 36 & $76.6\% \pm 2.0\%$ & 26 \\
		\hline
		\T\B Method 4 & $95.4\% \pm 1.0\%$ & 33 & $94.7\% \pm 1.0\%$ & 27 \\
		\hline
		\multicolumn{1}{c}{}\\
		\multicolumn{5}{c}{Approach 2}\\
		\hline
		& Accuracy &  Number of Breaks & Accuracy & Number of Breaks\\
		\hline
		\T\B Proposed Method &  $\mathbf{97.0\% \pm 0.8\%}$ & 0 &  $\mathbf{95.7\% \pm 0.9\%}$ & 0  \\
		\hline
		\T\B Method 2 & $88.3\% \pm 1.5\%$ & 6 & $93.5\% \pm 1.1\%$ & 0 \\
		\hline
		\T\B Method 3 & $66.1\% \pm 2.2\%$ & 36 & $76.5\% \pm 2.0\%$ & 11 \\
		\hline
		\T\B Method 4 & $86.9\% \pm 1.5\%$ & 47 & $91.9\% \pm 1.3\%$ & 34\\
		\hline
		\multicolumn{1}{c}{}\\
		\multicolumn{5}{c}{Approach 3}\\
		\hline
		&Accuracy & Number of Breaks & Accuracy & Number of Breaks\\
		\hline
		\T\B Proposed Method &  $96.9\% \pm 0.8\%$ & 0 &  $95.7\% \pm 0.9\%$ & 0  \\
		\hline
		\T\B Method 2 & $\mathbf{97.8\% \pm 0.7\%}$ & 9 & $\mathbf{96.7\% \pm 0.8\%}$ & 0 \\
		\hline
		\T\B Method 3 & $70.6\% \pm 2.1\%$ & 146 & $72.2\% \pm 2.1\%$ & 120\\
		\hline
		\T\B Method 4 & $95.5\% \pm 0.9\%$ & 160 & $94.3\% \pm 1.1\%$& 130\\
		\hline
	\end{tabular}}
	\caption{Experiment 2 results using a parsed version of DS 1 and DS2, where off road positions are excluded. The columns have the same interpretations as Table \ref{Table:Experm1Result}.}
	\label{Table:Experm2Result}
\end{table}

Since Method 5 does not use the Viterbi algorithm, its results are reported separately.
Table \ref{Table:Method5Result} presents the results of using Method 5 for both datasets, using three different values of its parameter $d$: $1$, $3$, and $5$. 
At each time instant, Method 5 selects the road segment as described in Section \ref{sec:LD_model}.
When the vehicle is not on the road, Method 5 will still report a closest lane; herein these instances are counted as incorrect for evaluating accuracy in the tabulation of results. 
Method 5 is able to achieve high accuracy in Experiment 2, when all off-road positions are removed.
Method 5 is also able to produce the highest accuracy with $d = 1$ in Experiment 2; however, 
because it only utilizes the estimated position at a single time to calculate the most probable lane, Method 5 will be the method most negatively affected by \ac{GNSS} inaccuracies, which explains why it is not consistently the best performing method.

\begin{table}
	\centering
	\resizebox{\columnwidth}{!}{
		\begin{tabular}{|c |c | c|}
			\multicolumn{3}{c}{Experiment 1 Results}\\
			\hline
			Observations used & DS 1 Accuracy & DS 2 Accuracy\\
			\hline
			 d = 1 &  $89.0\% \pm 1.4\%$ & $94.5\% \pm 1.0\%$ \\
			\hline
			 d = 3 &  $85.4\% \pm 1.5\%$ &  $91.1\% \pm 1.3\%$  \\
			\hline
			 d = 5 & $81.1\% \pm 1.7\%$ & $87.5\% \pm 1.5\%$ \\
			\hline
			\multicolumn{3}{c}{}\\
			\multicolumn{3}{c}{Experiment 2 Results}\\
			\hline
			Observations used & DS 1 Accuracy & DS 2 Accuracy\\
			\hline
			d = 1 &  $98.6\% \pm 0.5\%$ & $97.7\% \pm 0.7\%$ \\
			\hline
			 d = 3 &  $92.7\% \pm 1.2\%$ &  $94.6\% \pm 1.0\%$  \\
			\hline
			 d = 5 & $88.8\% \pm 1.4\%$ & $89.7\% \pm 1.4\%$ \\
			\hline
	\end{tabular}}
	\caption{Method 5 results.
	}
	\label{Table:Method5Result}
\end{table}

\section{Conclusion} \label{sec:Conc}
\noindent
This  paper  introduced a new lane determination approach that determines the  data-dependent  \ac{HMM} emission and transition model parameters by processing navigation system data relative to the road/lane geometry.
This method achieves enhanced reliability and accuracy because it accounts for the continuously changing probabilities as the vehicle travels along the road.
Two versions of a real-time Viterbi algorithm variation are also proposed as alternatives that have fixed computational requirements.

\subsection{Experimental Results}
The experiment results show that the data-dependent \ac{HMM} approach that relies on the lane geometry and vehicle state estimates, to calculate the \ac{HMM} transition and emission probabilities,
provides an increase in accuracy and reliability compared to existing lane determination methods in the literature.
Moreover, the proposed model does not require any training or parameter tuning, which improves generality.
The proposed model also provides a method of dealing with situations where the vehicle is not in a lane by introducing a no lane state with its own emission and transition probabilities.
Thus, removing the need for post-processing and removal of observations to deal with possible HMM breaks.

\subsection{Future Work}
The proposed model has only been tested using GNSS data to generate the set of observations $\mathcal{O}$.
Future work could explore including additional sensor and vehicle information.
These modifications will require an extension to the data-dependent \ac{HMM},
after which, the Viterbi Algorithm performance should improve further.\

\section{Acknowledgment} \label{sec:Acknowledge}
 
This work was partially funded by support from the U.S. Department of Energy EcoCAR EV Challenge, a gift from Mitsubishi Electric Research Laboratories (MERL), the National Center for Sustainable Transportation (NCST), and the UCR KA Endowment.


\appendix
The following appendix describes concepts important to the paper that are too detailed for the main body.

\subsection{Frame Definitions} \label{sec:frame}
Throughout the paper, the vehicle state vector $\BoldX_k= [\Boldp_k,\Boldv_k]^\top$ is represented in various frames-of-reference where ${\Boldp}_k$ and ${\Boldv}_k$ represent the position and velocity of the vehicle.
The three frames used in the paper consist of \ac{ECEF} (e), intersection (g), and lane ($\ell$) frame-of-reference.

\begin{description}
\item{\bf ECEF Frame ($e$)}
Every $T$ seconds, the \ac{GNSS} receiver provides position measurements ${}^e\tilde\Boldp \in \mathbb{R}^3$ in \ac{ECEF} frame.
The paper only utilizes Cartesian coordinates because it facilitates vector mathematics and simplifications. 
If a sensor or device communicates geodetic coordinates, the measurements will be converted to the equivalent Cartesian coordinates using the methods in Section 2.3.3 of \cite{farrell2008aided}.

\item{\bf Intersection Frame ($g$)}
The origin of the intersection frame $(g)$ is located at the point $\BoldT_{eg}$ with ECEF coordinates ${}^e\BoldT_{eg}$, which are known.
The axes of $g$-frame point in the north, east, and down directions. 

The transformation of ECEF Cartesian coordinates ${}^e\Boldp$ from $e$-frame to $g$-frame is described in Section 2.4.2 of \cite{farrell2008aided} as:
\begin{align}\label{eqn:Transf_e2g}
	{}^g\Boldp &= {}^g_e\BoldR \, \left({}^e\Boldp  - {}^e\BoldT_{eg} \right),
\end{align}
where the matrix ${}^g_e\BoldR$ rotates vectors from the e-frame to g-frame.
The rotation matrix ${}^g_e\BoldR$ is computed based on the origin ${}^e\BoldT_{eg}$, as discussed in Section 2.5.2 of \cite{farrell2008aided}.

\item{\bf Lane Frame ($\ell$)}
For lane segment $m$ of lane $i$, the origin is at the location with geodetic coordinates ${}^g\Boldp_{(v,m)}^i$.

The transformation of the $g$-frame position ${}^g\Boldp$ to $\ell$-frame for lane segment $m$ of the $i$-th lane is 
\begin{equation}
	{}^\ell_i\Boldp = {}^\ell_g\BoldR_{im} ({}^g\Boldp - {}^g\Boldp_{(v,m)}^i), \label{eqn:PointRotate}
\end{equation}
where the rotation matrix ${}^\ell_g\BoldR_{im}$ is defined as
\begin{equation}
	{}^\ell_g\BoldR_{im}  = 
	\begin{bmatrix}
		\cos \alpha & -\sin \alpha\\
		\sin \alpha & \cos \alpha\\
	\end{bmatrix}.\nonumber
\end{equation}
The angle $\alpha$ is defined as the rotation around the down vector to align the geodetic $\Bolde$-axis with the lane $\Bolds$-axis.
\end{description}

\noindent
When a covariance matrix is known in one frame and needed in another, it can be computed using the rotation matrix. 
For example, if ${}^g\BoldC_{\Boldp_k}$ is available and ${}^\ell\BoldC_{\Boldp_k}$ is required
\begin{equation}\label{eqn:CovMatrix}
	{}^\ell\BoldC_{\Boldp_k} = {}^\ell_g\BoldR_{im}~ {}^g\BoldC_{\Boldp_k} ~{}^\ell_g\BoldR_{im}^\top.
\end{equation}

\label{sec:Apndx}

\bibliographystyle{IEEEtran}
\bibliography{LD_HMM.bib}

\printacronyms[]
\vfill 
	
\begin{IEEEbiographynophoto}{Mike Stas}
	received the B.S and M.S degrees in electrical engineering from University of California, Riverside.
	He is pursuing his Ph.D degree in Electrical Engineering at the University of California, Riverside, USA. His research interests include Connected and Automated Vehicle (CAV) systems, GNSS, GNSS spoofing detection and its effects on CAV systems, aided inertial navigation, and state estimation.
\end{IEEEbiographynophoto}
\begin{IEEEbiographynophoto}{Wang Hu}
	received the B.Eng. degree in electronics and information engineering from North China Electric Power University, Beijing, China. He is pursuing his Ph.D. degree in Electrical and Computer Engineering at the University of California, Riverside, CA, USA. His research interests include automated driving, GNSS, aided inertial navigation, and state estimation.
\end{IEEEbiographynophoto}
\begin{IEEEbiographynophoto}{Jay Farrell}
	is the KA Endowed Professor of ECE at the University of California, Riverside. He has served as General Chair of the 2012 IEEE Conference on Decision and Control, President of the IEEE CSS, and President of the American Automatic Control Council. At Draper Lab, Farrell received the Engineering VP’s Best Technical Publication Award in 1990, and Recognition Awards for Outstanding Performance and Achievement in 1991 and 1993. He is a Fellow of the IEEE, AAAS, and IFAC and author of 3 books and over 250 technical publications.
\end{IEEEbiographynophoto}
\end{document}